%% file: main.tex
\documentclass{article}
\usepackage[nonatbib,final]{neurips_2020}

\usepackage[utf8]{inputenc} % allow utf-8 input
\usepackage[T1]{fontenc}    % use 8-bit T1 fonts
\usepackage{hyperref}       % hyperlinks
\usepackage{url}            % simple URL typesetting
\usepackage{booktabs}       % professional-quality tables
\usepackage{amsfonts}       % blackboard math symbols
\usepackage{nicefrac}       % compact symbols for 1/2, etc.
\usepackage{microtype}      % microtypography
\usepackage{xcolor}

\usepackage{graphicx}
\usepackage{subcaption}
\RequirePackage{amssymb,amsmath,amsthm,amsfonts}

\renewcommand{\leq}{\leqslant}
\usepackage[capitalize]{cleveref}

\title{Backdoor Attacks on Federated Meta-Learning}
\author{%
  Chien-Lun Chen\\
  Department of Electrical Engineering\\
  University of Southern California\\
  Los Angeles, USA\\
  \texttt{chienlun@usc.edu}
  \And
  Leana Golubchik {\normalfont and} Marco Paolieri\\
  Department of Computer Science\\
  University of Southern California\\
  Los Angeles, USA\\
  \texttt{\{leana,paolieri\}@usc.edu}
}

\begin{document}

\maketitle

\input{tex/abstract}
\input{tex/intro}
\input{tex/problem}
\input{tex/backdoor}
\input{tex/bd_result}
\input{tex/defense}
\input{tex/related}
\input{tex/conclusions}

\newpage
\section*{Acknowledgements}
This material is based upon work supported by the National Science Foundation under grants number CNS-1816887 and CCF-1763747.

\section*{Broader Impact}

%Authors are required to include a statement of the broader impact of their work, including its ethical aspects and future societal consequences.
%Authors should discuss both positive and negative outcomes, if any. For instance, authors should discuss a) who may benefit from this research, b) who may be put at disadvantage from this research, c) what are the consequences of failure of the system, and d) whether the task/method leverages biases in the data. If authors believe this is not applicable to them, authors can simply state this.

The context for this paper is Federated Learning (FL), a framework designed to allow users with insufficient but confidential data to jointly train machine learning models while preserving privacy.
For example, a single hospital, clinic, or public health agency may not have sufficient data to train powerful machine learning models providing doctors with diagnostic aid (e.g., for medical images such as X-rays or fMRI images): FL allows multiple of these entities to jointly train more accurate machine learning models without sharing patients' private data. Unfortunately, FL is known to be vulnerable to poisoning backdoor attacks, where a malicious user of the federation can control the decisions of the model trained jointly with benign users. In the context of diagnostic medical aid, backdoor attacks to FL could mislead doctors into making incorrect diagnoses, with life threatening risks; similarly, consequences of backdoor attacks could be disastrous for many applications in healthcare, transportation, finance. Although defense mechanisms have been proposed in the literature, state-of-the-art solutions against poisoning backdoor attacks rely on \emph{third parties to examine user updates to the model}, violating the fundamental {\em privacy} motivation of FL. We believe that an effective \emph{user-end defense mechanism} can guard against backdoor attacks while preventing unexpected abuse due to privacy leaks. Thus the broader impact of our proposed approach is that it can prevent attacks on machine learning models that are developed jointly by multiple entities as well as privacy-related abuse. Allowing multiple entities to jointly develop machine learning models is critical to the broader impact of machine learning applications in settings (e.g., healthcare) where data is scarce and sensitive.

% \nocite{refToInclude}
\bibliography{refs}
\bibliographystyle{ieeetr}

% \newpage
\input{tex/appendix}

\end{document}

%% file: tex/abstract.tex
% -*- ispell-local-dictionary: "american"; TeX-master: "../main.tex"; -*-

\begin{abstract}
Federated learning allows multiple users to collaboratively train a shared classification model while preserving data privacy. This approach, where model updates are aggregated by a central server, was shown to be vulnerable to \emph{poisoning backdoor attacks}: a malicious user can alter the shared model to arbitrarily classify specific inputs from a given class.
In this paper, we analyze the effects of backdoor attacks on federated \textit{meta-learning}, where users train a model that can be adapted to different sets of output classes using only a few  examples.
While the ability to adapt could, in principle, make federated learning frameworks more robust to backdoor attacks (when new training examples are benign), we find that even 1-shot~attacks can be very successful and persist after additional training.
To address these vulnerabilities, we propose a defense mechanism inspired by \emph{matching networks}, where the class of an input is predicted from the similarity of its features with a \emph{support set} of labeled examples. By removing the decision logic from the model shared with the federation, success and persistence of backdoor attacks are greatly reduced.
\end{abstract}

%% file: tex/intro.tex
% -*- ispell-local-dictionary: "american"; TeX-master: "../main.tex"; -*-

\iffalse
COMMENTS FROM MARCO
===================

I removed the ``machine learning has achieved tremendous success in...'' because this is ICML. It's better to start from ``Federated Learning.''

NEW INTRO OUTLINE
=================

Federated Learning
- train from multiple datasets without sharing data
- aggregation at the server does not need to see the updates
- problems:
  - backdoor attacks
  - same classification task for all users

Federated META-Learning
- train to learn an initialization that can be adapted
- different classification tasks for each user

Our contributions
- analyze the effect of backdoor attacks
- propose a defense mechanism
\fi

\section{Introduction}
\label{sec:intro}

\emph{Federated learning} \cite{McMahanMRHA17} allows multiple users to collaboratively train a shared prediction model without sharing their private data.
Similarly to the \emph{parameter server} architecture, model updates computed locally by each user (e.g., weight gradients in a neural network) are aggregated by a server that applies them and sends the updated model to the users.
User datasets are never shared, while the aggregation of multiple updates makes it difficult for an attacker in the federation to reconstruct training examples of another user.
Additional privacy threats can also be addressed in federated learning: for example, users can send encrypted updates that the server applies to an encrypted model \cite{BonawitzIKMMPRS17, PhongAHWM18}.

While the use of data from multiple users allows for improved prediction accuracy with respect to models trained separately, federated learning has been shown to be vulnerable to \emph{poisoning backdoor attacks} \cite{chen2017targeted,gu2017badnets}: a member of the federation can send model updates produced using malicious training examples where the output class indicates the presence of a hidden \emph{backdoor key}, rather than benign input features.
% Chien-Lun: The reason that I add "poisoning" to both Abstract and here is that we mention at page 2, line 50 "...after a single poisoned update", but we never mention our backdoor attack is one branch of poisoning attack until page 2, line 88.
%
This kind of attack can be successful after a single malicious update, and it is difficult to detect in practice because (1)~the attacker can introduce the backdoor with minimal accuracy reduction, and (2)~malicious updates can be masked within the distribution of benign ones \cite{BhagojiCMC19, bagdasaryan2018backdoor, baruch2019little}.
%%%{\bf from LG: not sure I understand what is meant by "within the distribution"?}

Another limitation of federated learning is due to the requirement that all users share the same output classes (e.g., the outputs of a neural network and their associated labels) and that, for each class, the distribution of input examples of different users be similar.
Recent approaches to \emph{gradient-based meta-learning} \cite{VinyalsBLKW16, FinnAL17, nichol2018first} provide a compelling alternative for federated scenarios: rather than training a model for a specific set of output classes, these methods try to learn model parameters that can be adapted very quickly to new classification tasks (with entirely different output classes) using only a few training examples (or ``shots'').
Meta-learning also allows users with different data distributions to jointly train a meta-model that they can adapt to their specific tasks. For example, in federated face recognition, each user trains a model using classification tasks from a distinct dataset (e.g., images of friends and relatives), but all users share the goal of training a meta-model to recognize human faces.

While the use of meta-learning in a federated setting and its privacy concerns were explored by previous work \cite{chen2018federated,li2019differentially}, \emph{the influence of backdoor attacks on federated meta-learning has not been investigated.}
Since meta-models have the ability to adapt to new classification tasks very quickly, it is unclear whether a backdoor attack can succeed and persist even with many users sharing {\em benign} updates of the meta-model and after fine-tuning the meta-model for a specific task with {\em benign} data.

This paper investigates backdoor attacks on federated meta-learning with the following contributions.

(1) We design a set of experiments to illustrate the effects of backdoor attacks under different scenarios.
Our results, presented in \cref{sec:bd_result}, show that backdoor attacks (triggering intentional misclassification) can be \emph{successful even after a single malicious update} from the attacker: in our Omniglot experiment, $80\%$ of backdoor examples are misclassified after a single poisoned update, regardless of whether they are from attacker's training set or from a separate validation set, while meta-testing accuracy is reduced by only $1\%$; in our mini-ImageNet experiment, $75\%$ of backdoor training examples and $50\%$ of backdoor validation examples are misclassified after a single malicious update, with $10\%$ reduction in meta-testing accuracy.
Moreover, \emph{the effects of the attack are persistent}, despite long meta-training after the attack (using only benign examples), or after fine-tuning of the meta-model by a benign user.
After many rounds of benign \emph{meta-training} on Omniglot, $50\%$ of backdoor examples are still misclassified as the attack target, in both training and validation datasets; on mini-ImageNet, success rates of the attack are reduced only from $75\%$ to $70\%$ (training) and from $50\%$ to $43\%$ (validation).
On both datasets, longer \emph{fine-tuning} of the meta-model by a benign user reduces the attack success rate by less than $10\%$.

(2) In \cref{sec:defense}, we propose a defense mechanism inspired by \emph{matching networks}~\cite{VinyalsBLKW16}, where the class of an input is predicted by a user from the similarity of its features with a \emph{support set} of examples.
By adopting this local decision mechanism, we reduce the success rate of backdoor attacks from as high as $90\%$ to less than $20\%$ (Omniglot training/validation and mini-ImageNet training) and from $50\%$ to $20\%$ (mini-ImageNet validation) in just a few iterations.
In \emph{contrast with other defense mechanisms} \cite{ShenTS16, BlanchardMGS17, SteinhardtKL17, QiaoV18, YinCRB18, xie2018generalized, Tran0M18, ChenSX18, MhamdiGR18, fung2018mitigating}, our method does not require any third-party to examine user updates, and it is thus compatible with secure aggregation of encrypted updates \cite{BonawitzIKMMPRS17, PhongAHWM18}.
%
%In addition, we make no assumptions about the distribution of training data at different users or the required fraction of benign users, thus {\em enabling the safe application of federated learning to a broader set of scenarios}.

%% file: tex/problem.tex
% -*- ispell-local-dictionary: "american"; TeX-master: "../main.tex"; -*-
\section{Backdoors in Federated Meta-Learning}
\label{sec:problem}

\paragraph{Federated Meta-Learning}
\emph{Federated learning} among $M$ users proceeds in rounds: in each Round~$t$, the server randomly selects $M_{\mathit{r}} \leq M$ users and transmits the shared model $\theta^t_G$ to them.
Each selected user $i$ initializes the local model $\theta_i^t$ to $\theta^t_G$, performs $E$ training steps, and then transmits the model update $\delta_i^t = \theta^t_i - \theta^t_G$ to the server.
As soon as $M_{\mathit{min}}$ of the $M_{\mathit{r}}$ updates are received, the server applies them to obtain the model for the next round $\theta^{t+1}_G = \theta^{t}_G + \sum_{i=1}^{M_{\mathit{min}}}\alpha_i\delta^{t}_i$, where $\alpha_i$ can be used to give more importance to the updates of users with larger datasets \cite{McMahanMRHA17}.

In federated \emph{meta-learning} \cite{chen2018federated}, training steps performed by each user on $\theta_i^t$ are designed to improve how well the model can be \emph{adapted to new classification tasks} (with different output classes), instead of improving its accuracy on a fixed task (with the same output classes for training and testing).
While second-order derivatives are needed to account for changes of gradients during the adaptation phase, first-order approximations have been proposed \cite{FinnAL17,nichol2018first}.
We adopt Reptile \cite{nichol2018first} for $K$-shot,~$N$-way meta-training: user~$i$ samples $B$~\emph{episodes} (a \emph{meta-batch}), each with a random set of $N$~output classes and $K$~training examples for each class. In each episode $j=1,\dots,B$, we use the $N K$ training examples to perform $e$ stochastic gradient descent~(SGD) steps (with \emph{inner} batch size~$b$ and learning rate~$\eta$) and to obtain a new model $\theta^{t,j}_i$ from $\theta^{t}_i$; models trained in different episodes are averaged to update $\theta^t_i$ as $\theta^t_i = (1-\epsilon)\theta^t_i + \frac{\epsilon}{B} \sum_{j=1}^B\theta^{t,j}_i$ (for some \emph{outer} learning rate~$\epsilon$).
To test a model after many rounds of $K$-shot,~$N$-way federated meta-training, the user generates new episodes, each with $N$~unseen classes and $K+1$ examples per class; for each episode, the shared model $\theta_G^t$ is fine-tuned with a few SGD steps on the first $K$ examples of each class and tested on the $N$ held-out examples.

\paragraph{Backdoor Attacks}
We consider backdoor attacks based on \emph{data poisoning} \cite{chen2017targeted, gu2017badnets, bagdasaryan2018backdoor, BhagojiCMC19}:
the attacker participates in the federation, applying the same meta-learning algorithm (Reptile) but using a poisoned dataset where examples from a \emph{backdoor class} are labeled as instances of a \emph{target class}; through model updates sent to the server, the attacker introduces changes in the shared model $\theta_G^t$ that persist after a benign user fine-tunes $\theta_G^t$ on a new classification task (with benign data).

For the attack to succeed, the target class must be present in the classification task of the user under attack, and images of the backdoor class must be used as inputs. Since classes are different in each meta-learning episode, the attacker can use multiple target and backdoor classes to increase success chances.
For example, in a face recognition problem, the attacker could collect online images $\mathcal X_T$ of a friend (the target class) of a member of the federation, and images $\mathcal X_B$ of a few impostors (the backdoor classes): in the training dataset of the attacker, examples of backdoor classes have the same label as images of a target class, so that the model learns to classify impostors as targets.

To ensure that the attack goes unnoticed, the attacker should also include valid data during training, so that the trained meta-model performs well on inputs that are not backdoor or target examples.
In particular, to generate an episode for $K$-shot, $N$-way meta-training, the attacker could pick $N-1$ random classes and always include the target class as the $N$-th model output: some of the $K$~examples of the target class are selected from $\mathcal X_T$, while others are selected from $\mathcal X_B$.
For \emph{attack-pattern backdoors}, the attacker can also add a special visual feature to the backdoor images $\mathcal X_B$, as a key to trigger the attack \cite{gu2017badnets, chen2017targeted}.
Similarly to poisoning attacks in federated learning \cite{BhagojiCMC19}, after many meta-training steps on the local model $\theta_a^t$, the attacker sends a ``boosted'' update to the parameter server: $\delta^{t}_a = \lambda(\theta^t_a-\theta^{t}_G)$, where $\lambda$ is the \emph{boosting factor} (to make it prevail over other updates).

%% file: tex/backdoor.tex
% -*- ispell-local-dictionary: "american"; TeX-master: "../main.tex"; -*-

%% file: tex/bd_result.tex
% -*- ispell-local-dictionary: "american"; TeX-master: "../main.tex"; -*-
\section{Effects of Backdoor Attacks}
\label{sec:bd_result}

In this section, we explore backdoor attacks on the \emph{Omniglot}~\cite{Lake1332} and \emph{mini-ImageNet} \cite{VinyalsBLKW16, sun2019mtl} datasets.

\noindent\textbf{Attack Evaluation.}\hspace{.5em}
We consider a federation of $M=4$ users, where user $i=1$ is the attacker and users $i=2,3,4$ are benign; at each round, the server selects 3~users and waits for all of their updates (i.e., $M_{\mathit{min}}=M_{\mathit{r}}=3$).
The meta-model is initially trained only by benign users, reaching state-of-the-art accuracy; then, the attacker is selected \emph{exactly once} (one-shot attack) and the poisoned update is boosted with $\lambda = 3$ \cite{bagdasaryan2018backdoor,BhagojiCMC19}.
To evaluate the effectiveness of the attack, we generate $K$-shot, $N$-way episodes from meta-training classes that always include the target class (with benign examples): after each fine-tuning iteration, we measure accuracy on testing examples of the episode (\emph{main-task accuracy}), as well as the percentage of poisoned backdoor examples labeled as the target (\emph{backdoor accuracy}); we separately evaluate backdoor accuracy on examples used by the attacker during training (\emph{attack training}) and on unseen examples (\emph{attack validation}). We also evaluate \emph{meta-testing accuracy} on other classes not used during meta-training.
Reported accuracy is averaged over $40$ episodes.

\begin{figure}[!t]
\centering
\begin{minipage}[t]{.48\textwidth}
\centering
\includegraphics[width=\textwidth]{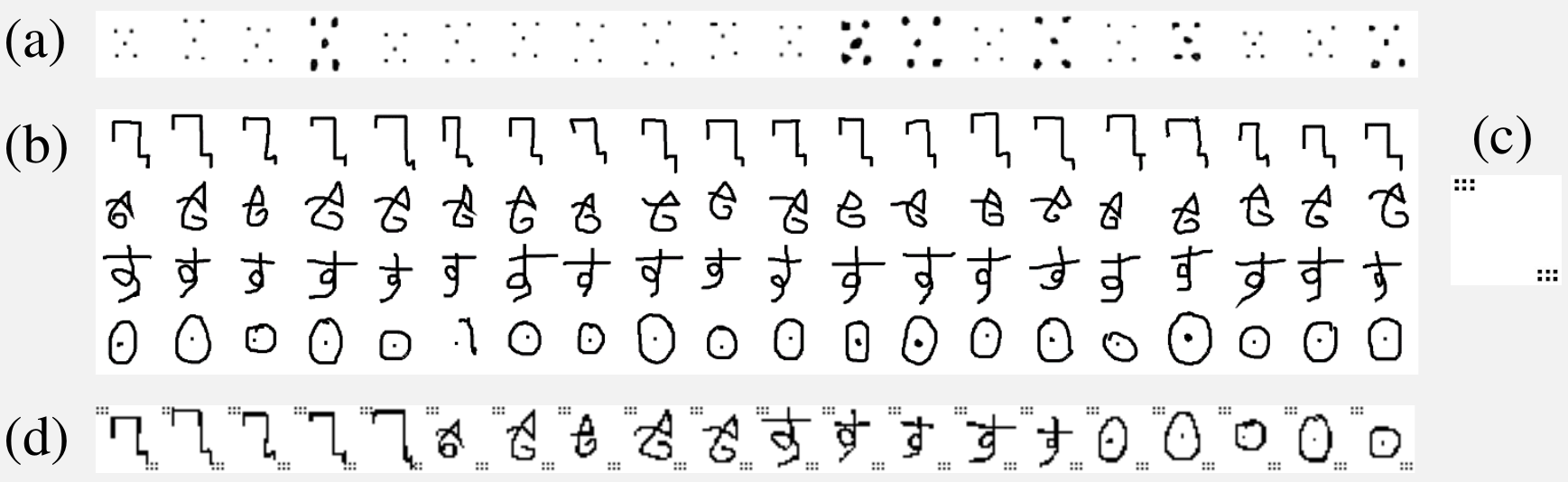}
\begin{minipage}[t]{\textwidth}
\caption{Backdoor attack on Omniglot: (a) target class, (b)  backdoor classes, (c) backdoor key, (d) attack training set}
\label{fig:omniglot}
\end{minipage}
\end{minipage}%
\hfill
\begin{minipage}[t]{.48\textwidth}
\centering
\includegraphics[width=\textwidth]{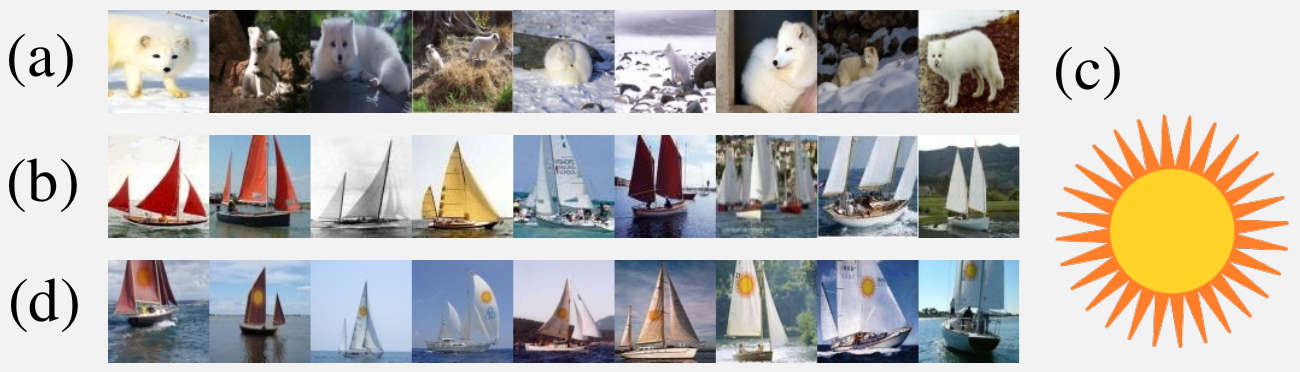}
\begin{minipage}[t]{\textwidth}
\caption{Backdoor attack on mini-ImageNet: (a) target class (arctic fox), (b) backdoor class (yawl), (c) backdoor key, (d) attack training set}
\label{fig:miniimagenet}
\end{minipage}
\end{minipage}
\vspace*{-4mm}
\end{figure}

\noindent\textbf{Omniglot.}\hspace{.5em}
This dataset consists of $1623$~character classes from $50$~alphabets, with $20$~examples per class. Similarly to \cite{nichol2018first, FinnAL17, SantoroBBWL16}, we resize images to $28\times28$ and augment classes $4\times$ using rotations: we use $1200$ classes for meta-training (split among the $4$~users) and $418$ for meta-testing; for each meta-training class, we hold out $5$ examples for validation.
We reserve $4$~backdoor classes and $1$~target class (\cref{fig:omniglot}a-b) for the attack: $10$~examples of each backdoor/target class are assigned to benign clients for training, while $5$ are edited to add a backdoor key (\cref{fig:omniglot}c-d) and used by the attacker.

\noindent\textbf{mini-ImageNet.}\hspace{.5em}
This dataset includes $100$ classes, each with $600$~examples ($84\times84$ color images). We use $64$~classes for meta-training (split among $4$~users) and $20$ classes for meta-testing, as in \cite{sun2019mtl}; for each meta-training class, we hold out $20$ validation examples.
We also reserve $1$~backdoor and $1$~target class (\cref{fig:miniimagenet}a-b): $480$ examples of each of these are split among benign clients for meta-training, while $100$ are used by the attacker as benign training examples. As attack and validation sets, we use $100$ and $50$ additional examples, respectively, adding a backdoor key as in \cref{fig:miniimagenet}c-d.

\noindent\textbf{Training Parameters.}\hspace{.5em}
All users run \emph{Reptile} on the same Conv4 model as in \cite{FinnAL17, nichol2018first}, a stack of $4$ modules ($3\times 3$ Conv filters with batchnorm and ReLU) followed by a fully-connected and a softmax layer; the modules have $64$ filters and $2 \times 2$ max-pooling.
We adopt the same parameters as in \cite{nichol2018first}: $5$-shot,~$5$-way meta-testing of a meta-model trained with $E = 1000$ episodes $10$-shot,~$5$-way (Omniglot) or $E = 100$ episodes $15$-shot,~$5$-way (mini-ImageNet) per round at each user, with meta-batch size $B = 5$ and outer learning rate $\epsilon = 0.1$; for each episode, we use $e = 10$ (meta-training) or $e=50$ (meta-testing) SGD steps, with inner batch size $b = 10$ and Adam optimizer ($\beta_1=0$, $\beta_2=0.999$, initial learning rate $\eta=0.001$).
The attacker trains for $E = 50000$ episodes and $50$ inner epochs (Omniglot), or $E = 150000$ and $1$ inner epoch (mini-ImageNet); backdoor and target examples $\mathcal X_B$ and $\mathcal X_T$ are always included by the attacker with $2$:$3$ (Omniglot) or $1$:$2$ (mini-ImageNet) ratio.

In our first set of experiments, benign users continue \emph{federated meta-training} after the attack.

\begin{figure}[t]
     \centering
     \begin{minipage}[t]{.48\textwidth}
     \centering
     \begin{subfigure}[hbt!]{\textwidth}
         \centering
         \includegraphics[width=\textwidth]{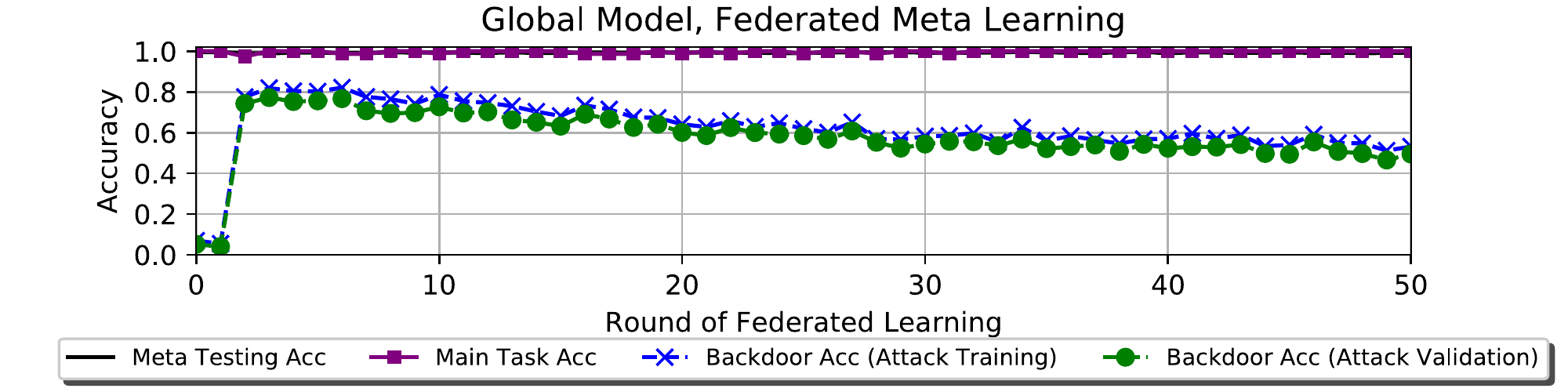}
         \vspace*{-6mm}
         \caption{Backdoor examples not used by benign users}
         \label{fig:001_1}
     \end{subfigure}
     \hfill
     \vspace*{1mm}
     \begin{subfigure}[hbt!]{\textwidth}
         \centering
         \includegraphics[width=\textwidth]{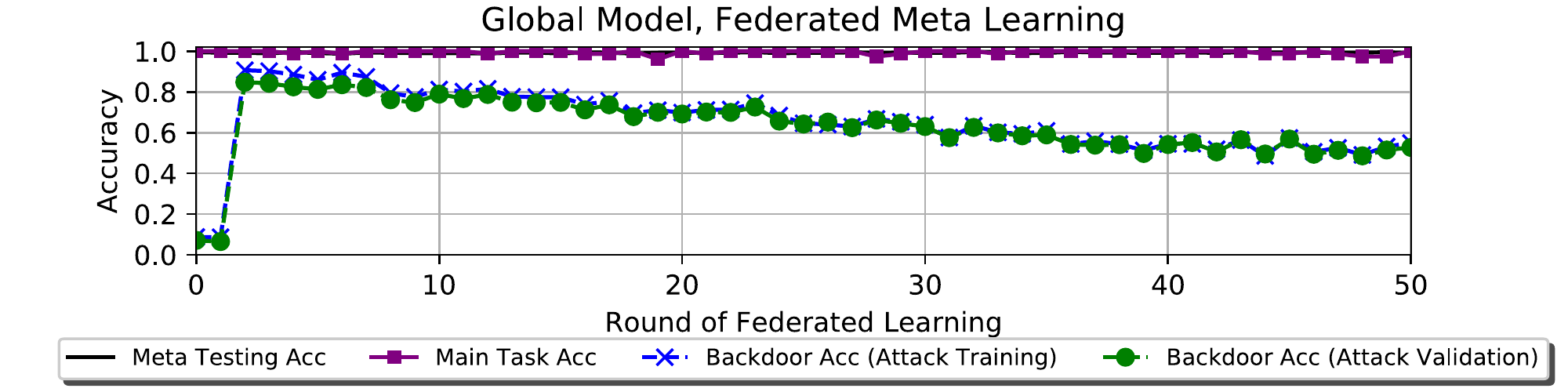}
         \vspace*{-6mm}
         \caption{Backdoor classes used in benign meta-training}
         \label{fig:001_2}
     \end{subfigure}
     \hfill
     \vspace*{1mm}
     \begin{subfigure}[hbt!]{\textwidth}
         \centering
         \includegraphics[width=\textwidth]{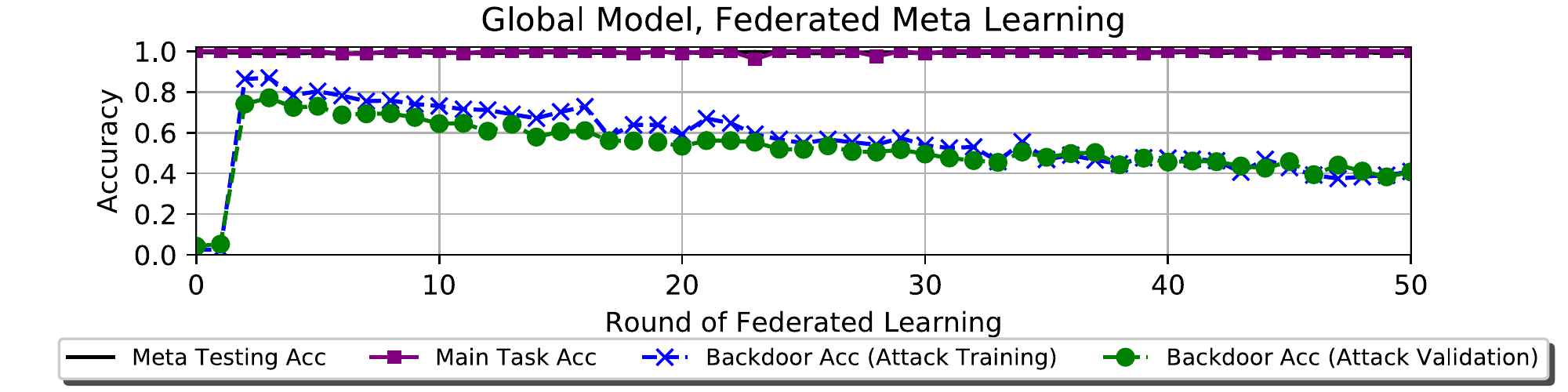}
         \vspace*{-6mm}
         \caption{Backdoor classes also used in benign fine-tuning}
         \label{fig:001_3}
     \end{subfigure}
     % \vspace*{-3mm}
     \caption{Benign meta-training after attacks on Omniglot }
     \vspace*{-3mm}
     \label{fig:001}
     \end{minipage}
     \hfill
     \begin{minipage}[t]{.48\textwidth}
     \centering
     \begin{subfigure}[hbt!]{\textwidth}
         \centering
         \includegraphics[width=\textwidth]{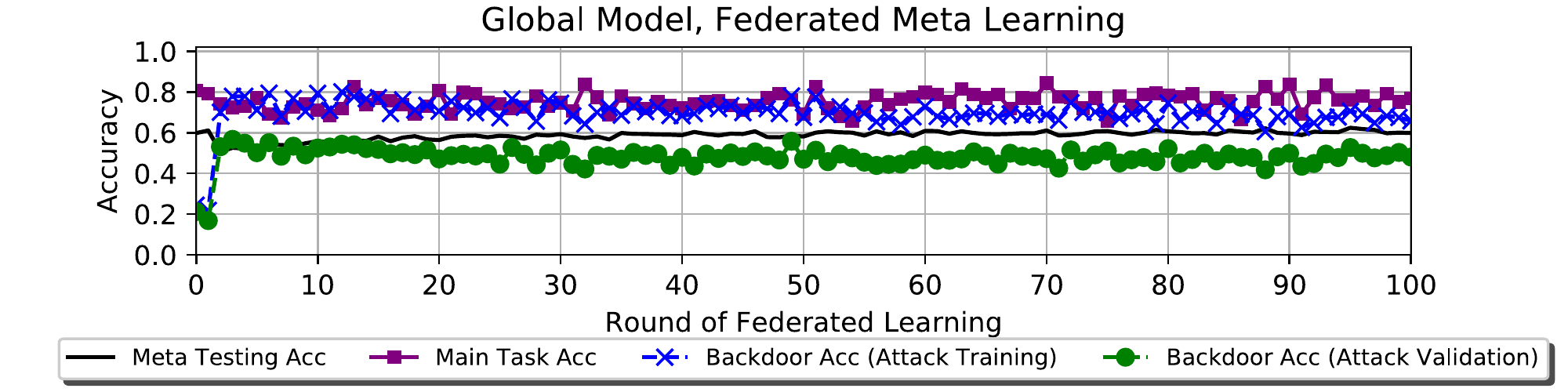}
         \vspace*{-6mm}
         \caption{Backdoor examples not used by benign users}
         \label{fig:001m_1}
     \end{subfigure}
     \hfill
     \vspace*{1mm}
     \begin{subfigure}[hbt!]{\textwidth}
         \centering
         \includegraphics[width=\textwidth]{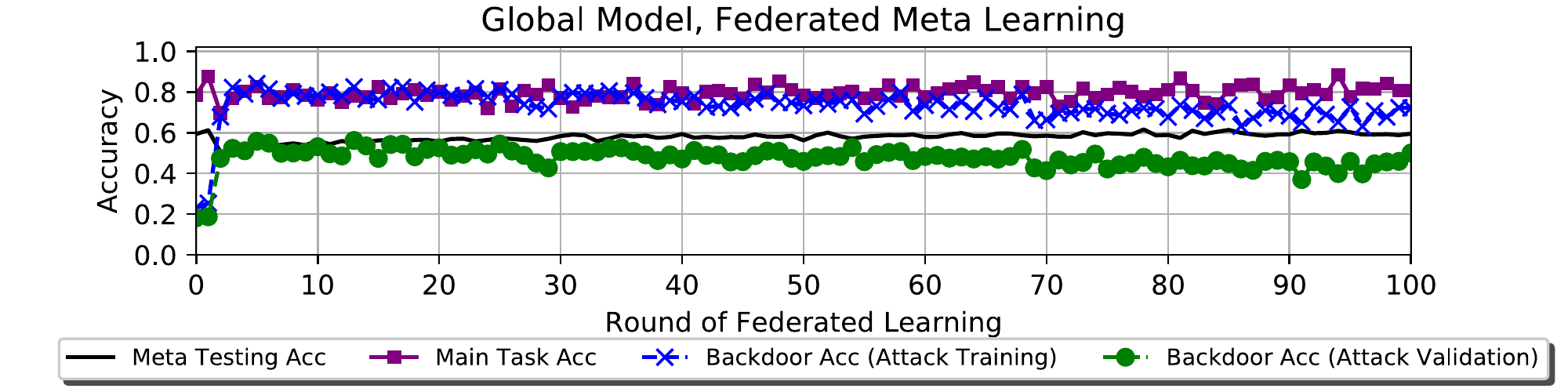}
         \vspace*{-6mm}
         \caption{Backdoor classes used in benign meta-training}
         \label{fig:001m_2}
     \end{subfigure}
     \hfill
     \vspace*{1mm}
     \begin{subfigure}[hbt!]{\textwidth}
         \centering
         \includegraphics[width=\textwidth]{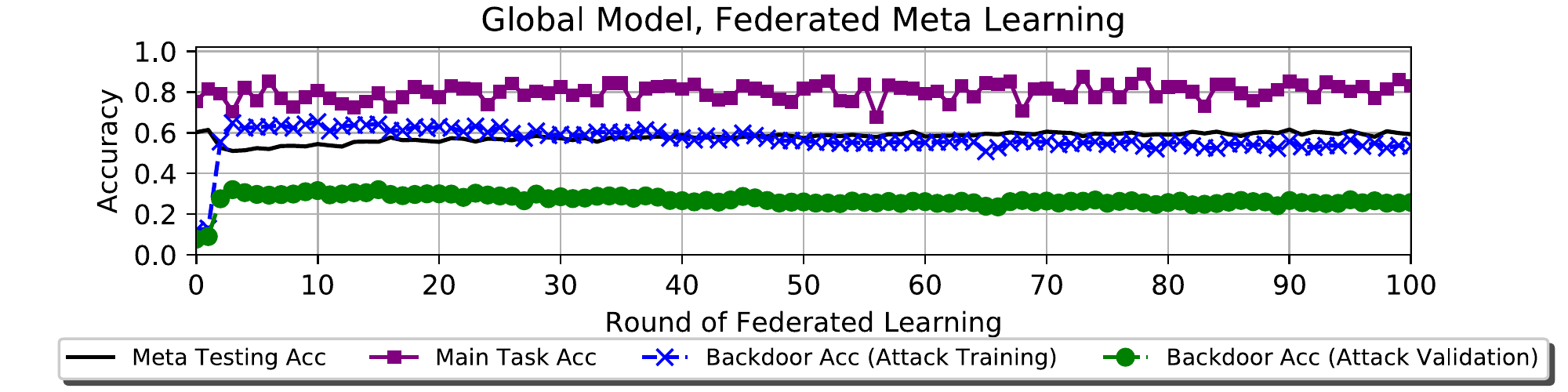}
         \vspace*{-6mm}
         \caption{Backdoor classes also used in benign fine-tuning}
         \label{fig:001m_3}
     \end{subfigure}
     % \vspace*{-3mm}
     \caption{Benign meta-training after attacks on mini-ImageNet}
     \vspace*{-3mm}
     \label{fig:001m}
     \end{minipage}
     \vspace*{-3mm}
\end{figure}

\noindent\textbf{Experiment 1(a).}\hspace{.5em}
First, we consider the case where initial meta-training by benign users does not include correctly-labeled examples of backdoor classes.
Results are in \cref{fig:001_1} (Omniglot) and \cref{fig:001m_1} (mini-ImageNet): before the attack (Round~0), meta-testing accuracy (black line) is above $99\%$ (Omniglot) or $60\%$ (mini-ImageNet); the attacker is selected in Round~1; then in Round~2 attack accuracy (classification of backdoor images as target class) reaches $78\%$ and $74\%$ on attacker's training dataset (blue line) and $77\%$ and $55\%$ on the held-out validation set (green line) for Omniglot and mini-ImageNet, respectively, while meta-testing accuracy on other classes remains above $98\%$ (Omniglot) or drops to $50\%$ (mini-ImageNet).
Even after 50 (Omniglot) and 100 (mini-ImageNet) rounds of additional meta-training by benign users, backdoor accuracy is still high ($50\%$ on both attack training/validation for Omniglot; $68\%$ / $48\%$ on attack training/validation for mini-ImageNet).

\noindent\textbf{Experiment 1(b).}\hspace{.5em}
Next, we consider the case where meta-training datasets of benign users include correctly-labeled images of backdoor classes during pre-training, so the meta-model should easily adapt to classifying them correctly.
Results are in \cref{fig:001_2} (Omniglot) and \cref{fig:001m_2} (mini-ImageNet): meta-testing accuracy is still above $98\%$ and $\approx 50\%$ after the attack for Omniglot and mini-ImageNet, respectively, while attack training/validation accuracy is close to $92\%$ / $ 83\%$ (Omniglot) and $76\%$ / $50\%$ (mini-ImageNet); after additional meta-training by benign users, attack training/validation accuracy is still $50\%$ / $50\%$ (50 rounds, Omniglot) and $69\%$ / $42\%$ (100 rounds, mini-ImageNet).

\noindent\textbf{Experiment 1(c).}\hspace{.5em}
Finally, we investigate the case where backdoor classes are present, with correct labels, \emph{also during fine-tuning} (at meta-testing) at benign users; this is particularly relevant since fine-tuning should adapt the meta-model to these examples.
Results are in \cref{fig:001_3,fig:001m_3}: after~the attack (Round~2), meta-testing accuracy is still greater than $98\%$ (Omniglot) and $50\%$ (mini-ImageNet); however, attack training/validation accuracy drops to $90\%$ / $75\%$ (Omniglot) and $65\%$ / $32\%$ (mini-ImageNet), and, after additional meta-training by benign users, further drops to $40\%$ / $40\%$  (50 rounds, Omniglot), and $55\%$ / $25\%$ (100 rounds, mini-ImageNet), lower than previous results.
%using benign backdoor examples in additional meta-training \emph{and} fine-tuning gradually reduces attack accuracy.

Overall, we observe that backdoor attacks are: (1)~more successful on the attack training set (especially for mini-ImageNet), as expected; (2)~similarly successful when benign users use correctly-labeled backdoor images for meta-training; (3)~considerably less successful when fine-tuning also includes correctly-labeled backdoor images.
Nonetheless, \emph{it does not appear possible to rely only on additional meta-training to remove backdoor attacks.}
In our next set of experiments, we explore whether \emph{additional fine-tuning} (in meta-testing episodes) can remove the attack by leveraging the ability of meta-models to quickly adapt to a specific task.
We stop meta-training after the one-shot attack (Round~2) and start fine-tuning at each benign user using only correctly labeled examples.

\begin{figure}[t]
     \centering
     \begin{minipage}[t]{.48\textwidth}
     \centering
     \begin{subfigure}[hbt!]{\textwidth}
         \centering
         \includegraphics[width=\textwidth]{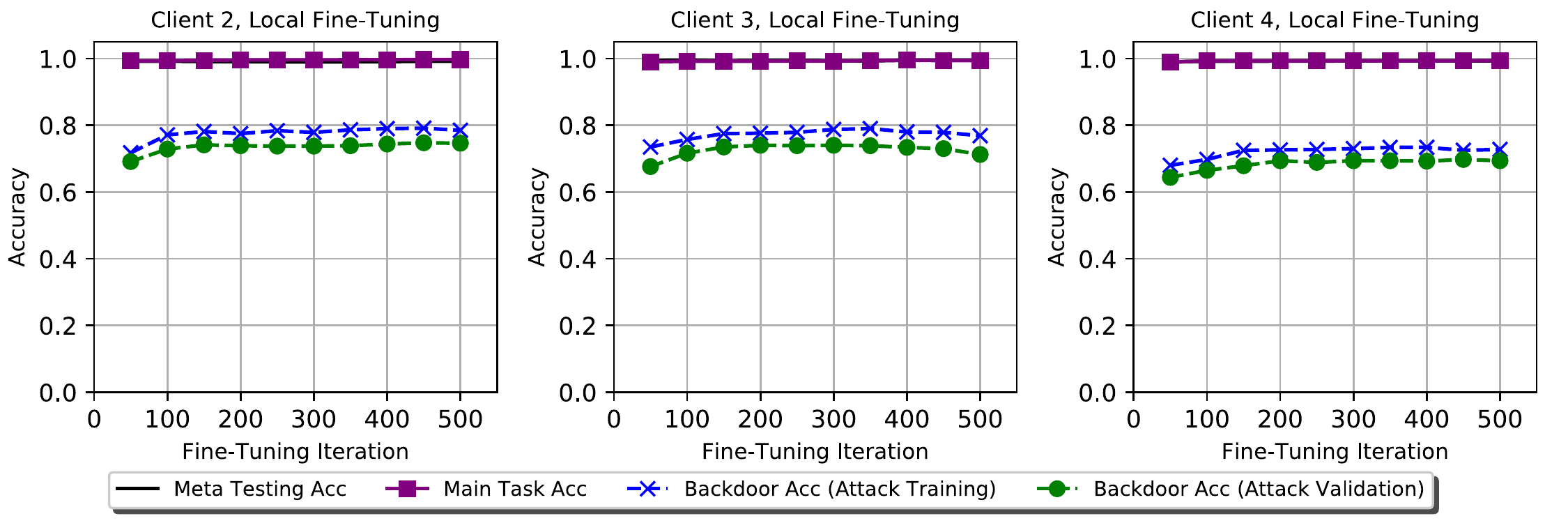}
         \vspace*{-6mm}
         \caption{Backdoor examples not used by benign users }
         \label{fig:002_1}
     \end{subfigure}
     \hfill
     \vspace*{1mm}
     \begin{subfigure}[hbt!]{\textwidth}
         \centering
         \includegraphics[width=\textwidth]{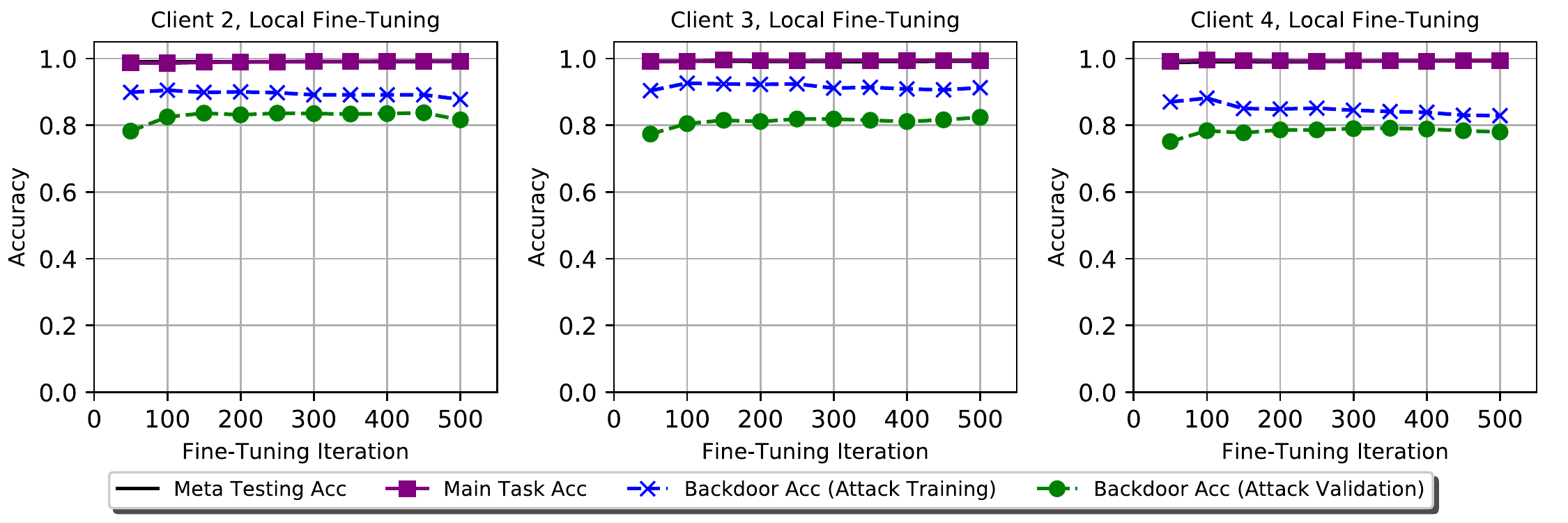}
         \vspace*{-6mm}
         \caption{Backdoor classes used in benign pre-training}
         \label{fig:002_2}
     \end{subfigure}
     \hfill
     \vspace*{1mm}
     \begin{subfigure}[hbt!]{\textwidth}
         \centering
         \includegraphics[width=\textwidth]{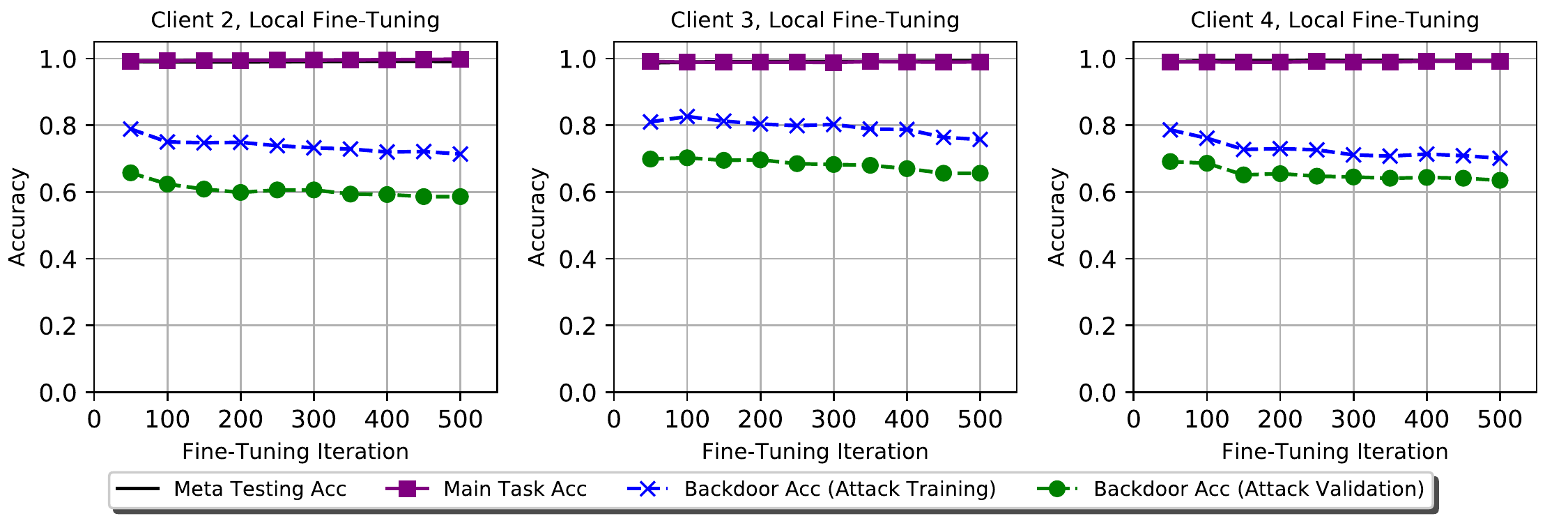}
         \vspace*{-6mm}
         \caption{Backdoor classes used also in benign fine-tuning}
         \label{fig:002_3}
     \end{subfigure}
     % \vspace*{-1mm}
     \caption{Benign fine-tuning ($\eta = 0.001$) after attacks on Omniglot}
     \vspace*{-3mm}
     \label{fig:002}
     \end{minipage}
     \hfill
     \begin{minipage}[t]{.48\textwidth}
     \centering
     \begin{subfigure}[hbt!]{\textwidth}
         \centering
         \includegraphics[width=\textwidth]{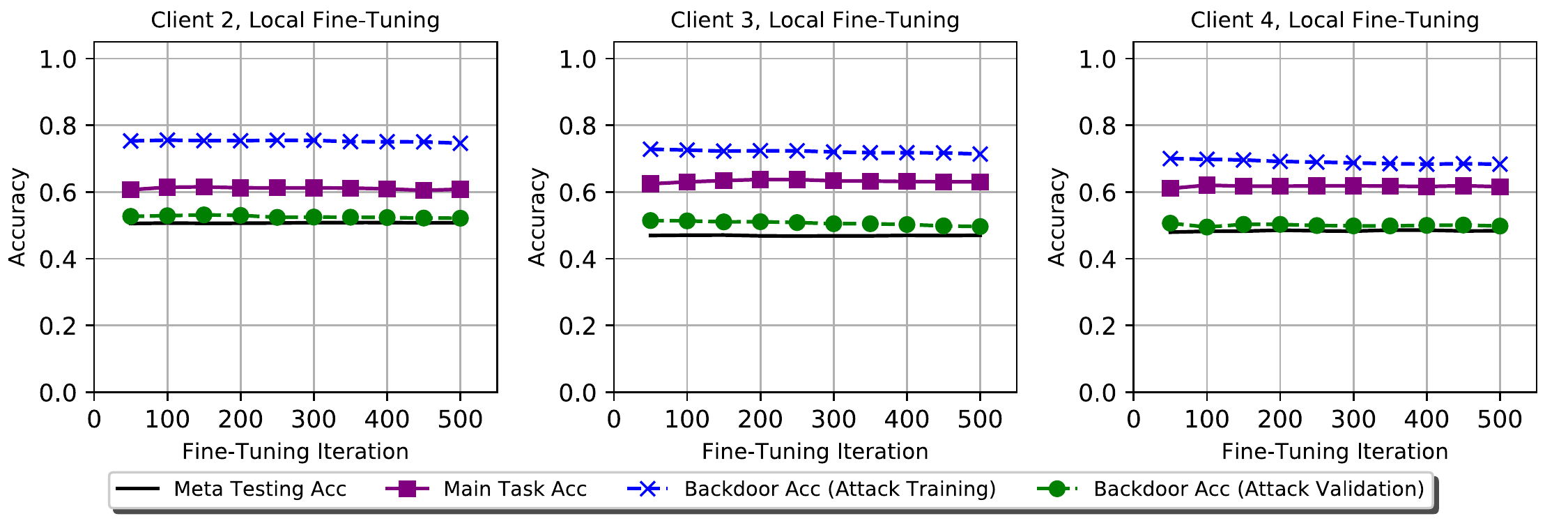}
         \vspace*{-6mm}
         \caption{Backdoor examples not used by benign users }
         \label{fig:002m_1}
     \end{subfigure}
     \hfill
     \vspace*{1mm}
     \begin{subfigure}[hbt!]{\textwidth}
         \centering
         \includegraphics[width=\textwidth]{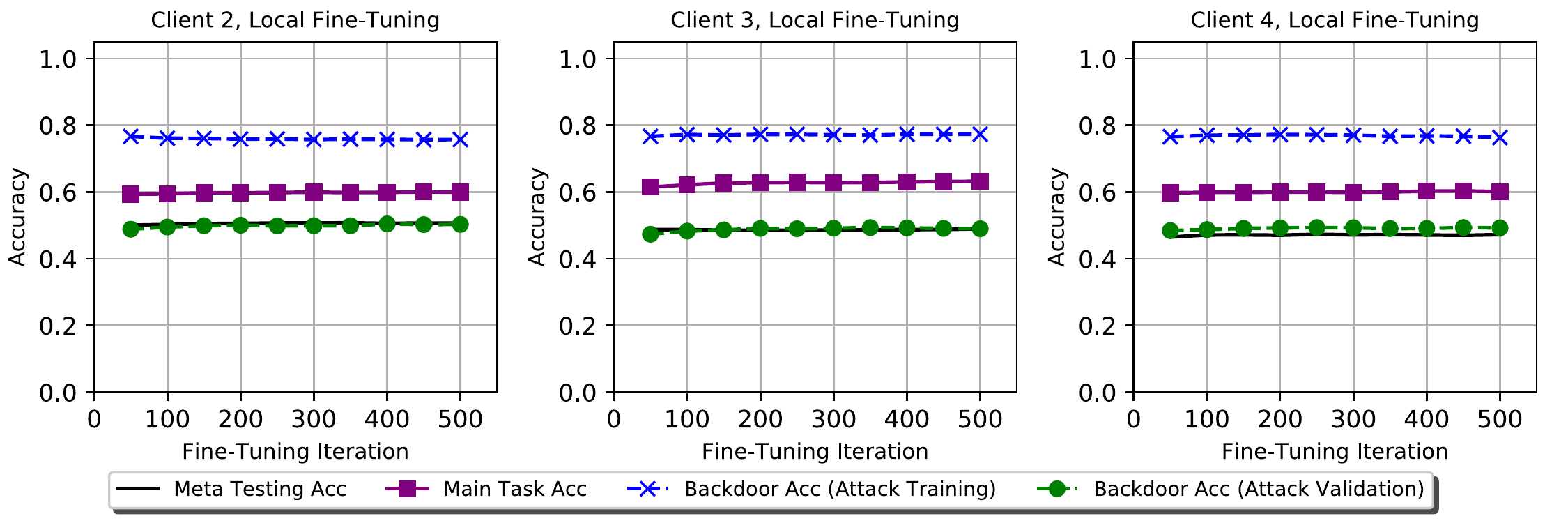}
         \vspace*{-6mm}
         \caption{Backdoor classes used in benign pre-training}
         \label{fig:002m_2}
     \end{subfigure}
     \hfill
     \vspace*{1mm}
     \begin{subfigure}[hbt!]{\textwidth}
         \centering
         \includegraphics[width=\textwidth]{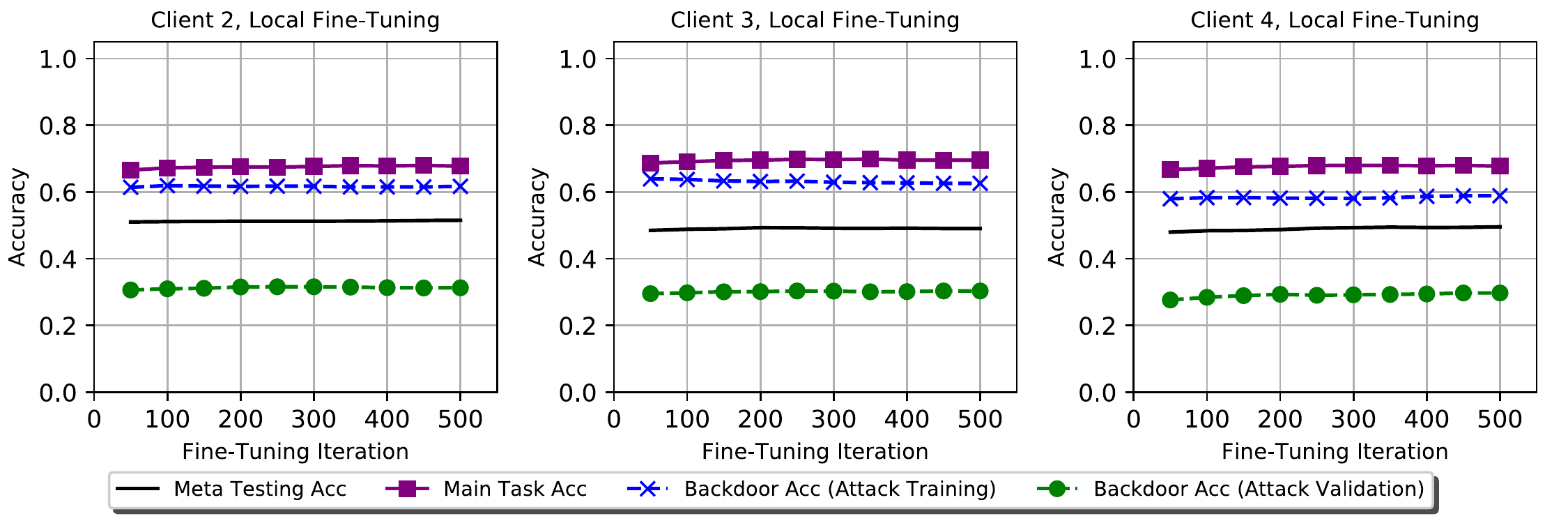}
         \vspace*{-6mm}
         \caption{Backdoor classes used also in benign fine-tuning}
         \label{fig:002m_3}
     \end{subfigure}
     % \vspace*{-1mm}
     \caption{Benign fine-tuning ($\eta = 0.001$) after attacks on mini-ImageNet}
     \vspace*{-3mm}
     \label{fig:002m}
     \end{minipage}
     \vspace*{-2mm}
\end{figure}

\noindent\textbf{Experiment 2.}\hspace{.5em}
We use the same learning rate $\eta=0.001$, but run $e=500$ ($10\times$ more) iterations of fine-tuning in Round 2 (right after the attack). 
%($10\times$ more than previous meta-testing). 
Results are in \cref{fig:002} (Omniglot) and \ref{fig:002m} (mini-ImageNet), with a column for each user and a row for each use case of correctly labeled backdoor examples: (a) not used, (b) used only during pre-training, (c) used also during fine-tuning.
% Chien-Lun: We forgot to mention that this is performed in Round 2, right after the attack.
%
\emph{Additional fine-tuning is also unsuccessful at removing the attack:} for Omniglot, both main-task accuracy (purple line) and meta-testing accuracy (black line) are above $99\%$ for all users. Backdoor accuracy is above $80\%$ for all users when backdoor classes are not present during fine-tuning (\cref{fig:002_1,fig:002_2});
when backdoor classes are present (\cref{fig:002_3}), attack accuracy drops slightly for all users, and it is gradually reduced during fine-tuning ($\approx 10\%$ after $500$ iterations).
For mini-ImageNet, when backdoor classes are not present during fine-tuning (\cref{fig:002m_1,fig:002m_2}), accuracy is $\approx 60\%$ (main-task) and $\approx 50\%$ (meta-testing) for all users. Backdoor accuracy for all users is $\approx 75\%$ (attack training) and $50\%$ (attack validation); however, when backdoor classes are present during fine-tuning (\cref{fig:002m_3}), main-task accuracy is improved by $5\%$ and attack accuracy is reduced by $20\%$ for all users.
From \cref{fig:002_3} and \cref{fig:002m_3}, we observe that the presence of backdoor classes has limited influence on attack accuracy.

%% file: tex/defense.tex
% -*- ispell-local-dictionary: "american"; TeX-master: "../main.tex"; -*-
\section{Matching Networks as a Defense Mechanism}
\label{sec:defense}

\begin{figure}[t]
     \centering
     \begin{minipage}[t]{.495\textwidth}
     \centering
     \begin{subfigure}[hbt!]{\textwidth}
         \centering
         \includegraphics[width=\textwidth]{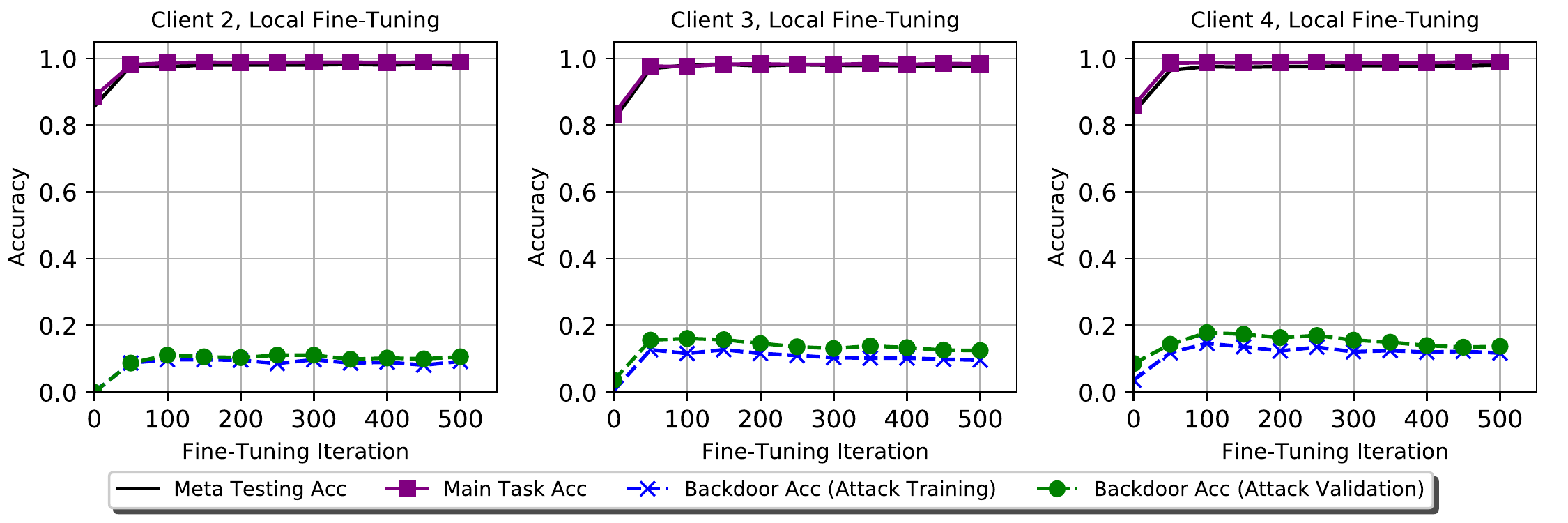}
         \vspace*{-6mm}
         \caption{Backdoor examples not used by benign users}
         \label{fig:005_1}
     \end{subfigure}
     \hfill
     \vspace*{1mm}
     \begin{subfigure}[hbt!]{\textwidth}
         \centering
         \includegraphics[width=\textwidth]{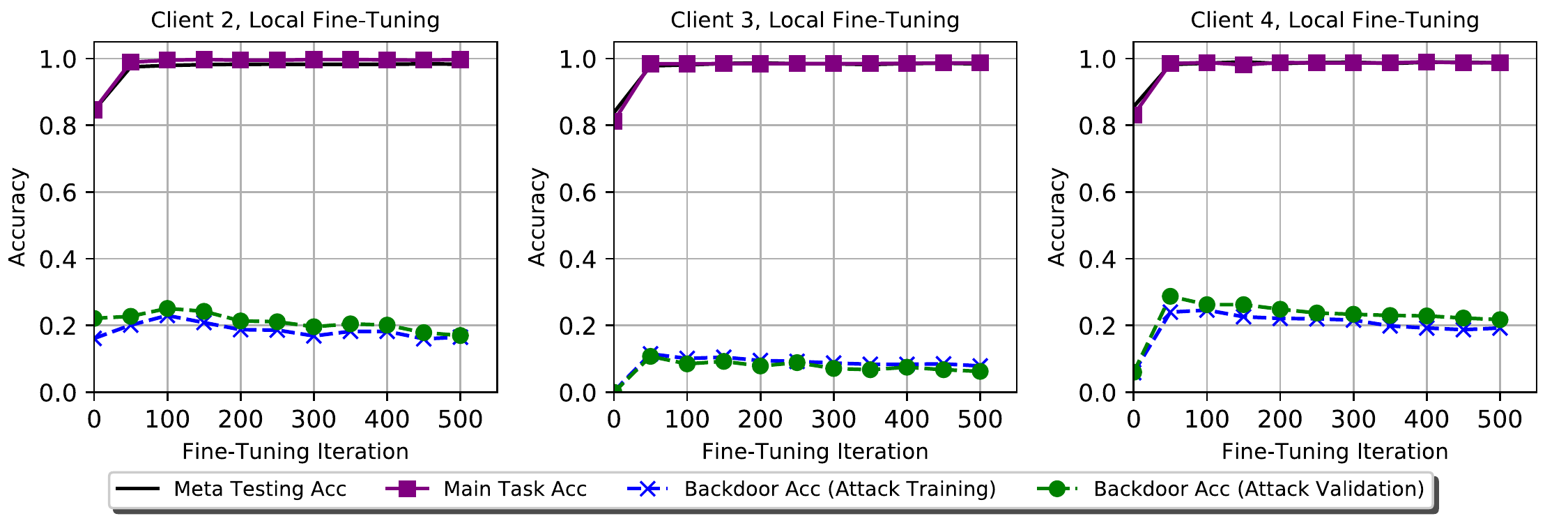}
         \vspace*{-6mm}
         \caption{Backdoor classes used in benign pre-training}
         \label{fig:005_2}
     \end{subfigure}
     \hfill
     \vspace*{1mm}
     \begin{subfigure}[hbt!]{\textwidth}
         \centering
         \includegraphics[width=\textwidth]{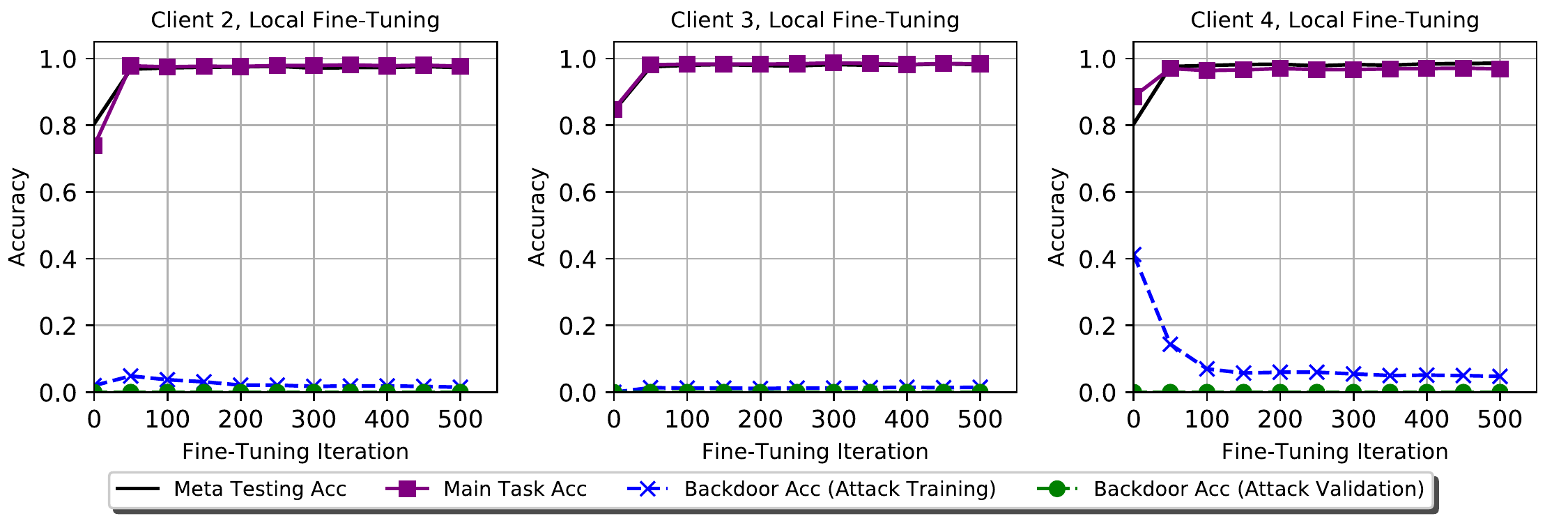}
         \vspace*{-6mm}
         \caption{Backdoor classes used also in benign fine-tuning}
         \label{fig:005_3}
     \end{subfigure}
     % \vspace*{-1mm}
     \caption{Benign fine-tuning of matching networks ($\eta = 0.001$) after attacks on Omniglot}
     \vspace*{-3mm}
     \label{fig:005}
     \end{minipage}
     \hfill
     \centering
     \begin{minipage}[t]{.495\textwidth}
     \centering
     \begin{subfigure}[hbt!]{\textwidth}
         \centering
         \includegraphics[width=\textwidth]{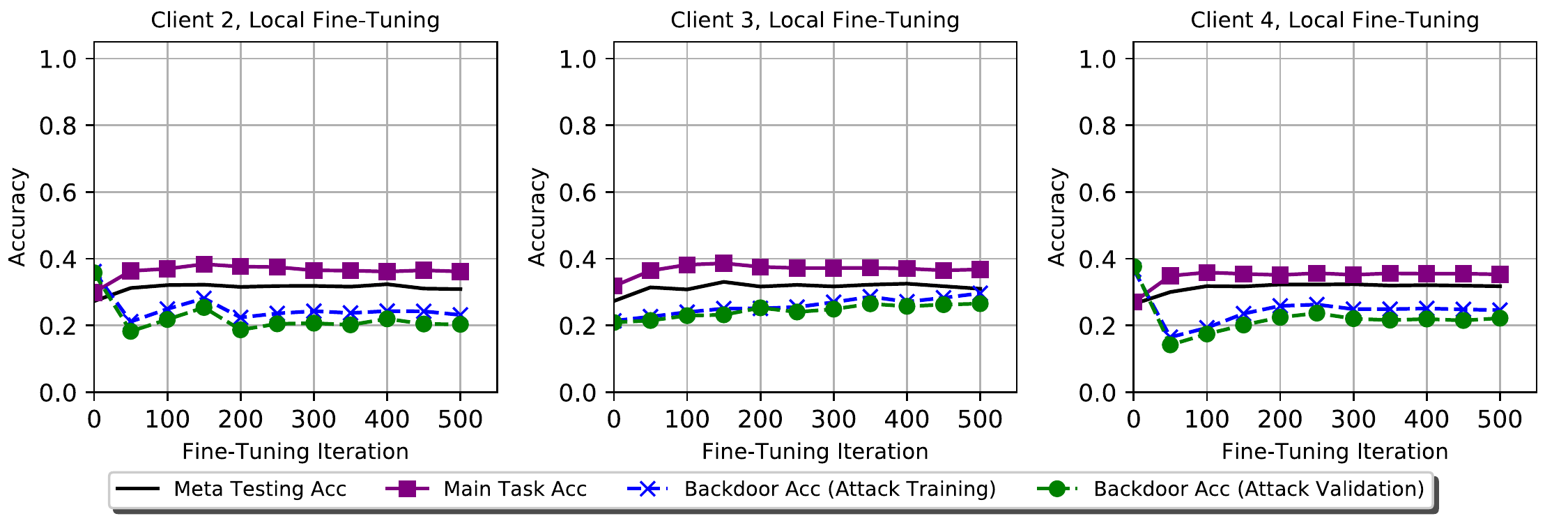}
         \vspace*{-6mm}
         \caption{Backdoor examples not used by benign users}
         \label{fig:005m_1}
     \end{subfigure}
     \hfill
     \vspace*{1mm}
     \begin{subfigure}[hbt!]{\textwidth}
         \centering
         \includegraphics[width=\textwidth]{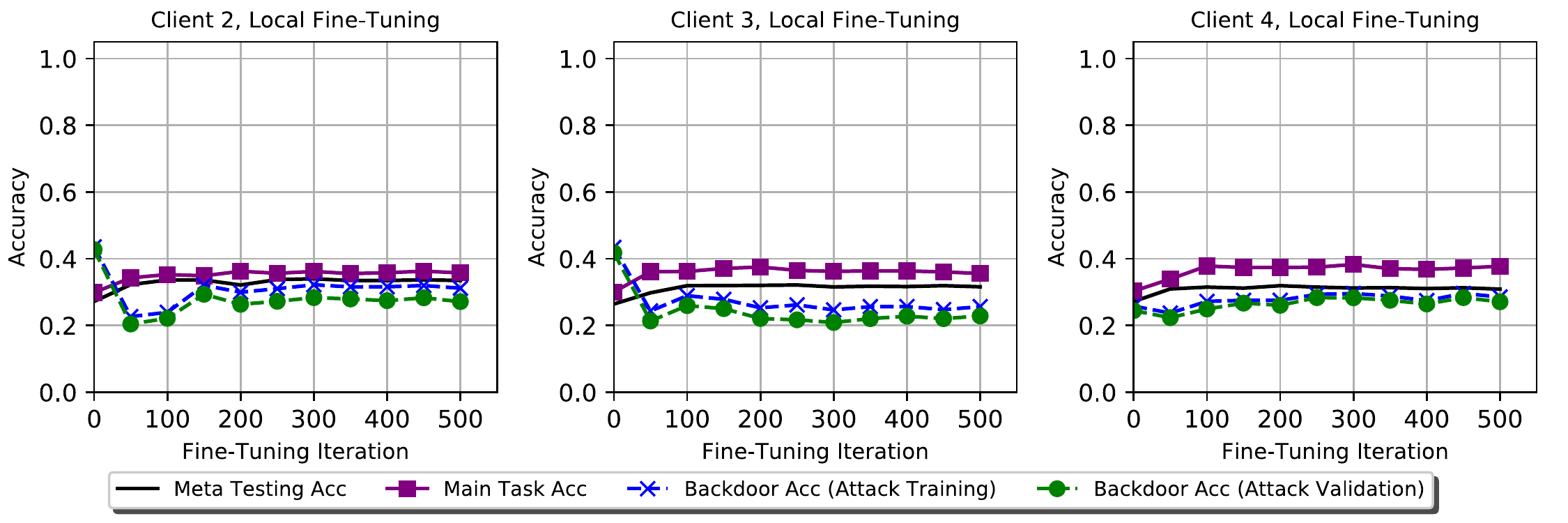}
         \vspace*{-6mm}
         \caption{Backdoor classes used in benign pre-training}
         \label{fig:005m_2}
     \end{subfigure}
     \hfill
     \vspace*{1mm}
     \begin{subfigure}[hbt!]{\textwidth}
         \centering
         \includegraphics[width=\textwidth]{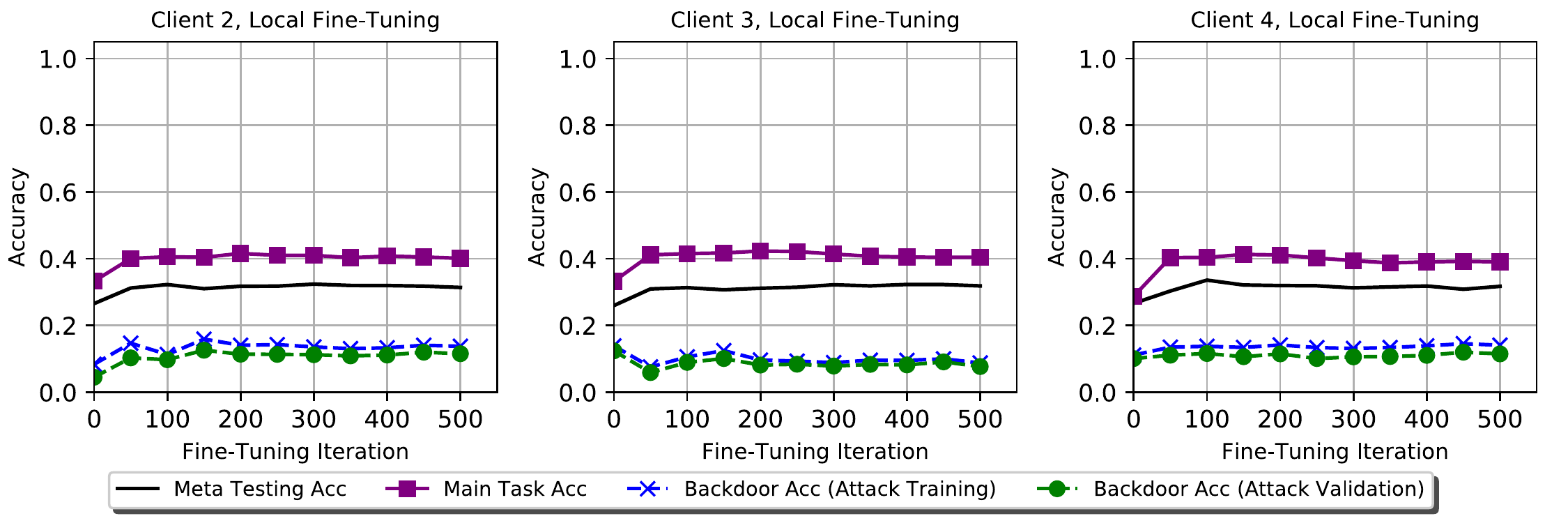}
         \vspace*{-6mm}
         \caption{Backdoor classes used also in benign fine-tuning}
         \label{fig:005m_3}
     \end{subfigure}
     % \vspace*{-1mm}
     \caption{Benign fine-tuning of matching networks ($\eta = 0.001$) after attacks on mini-ImageNet}
     \vspace*{-3mm}
     \label{fig:005m}
     \end{minipage}
     \vspace*{-2mm}
\end{figure}

Since defense mechanisms based on the analysis of updates received from users may violate privacy and are not compatible with secure update aggregation by the server, we propose a defense mechanism \emph{applied locally by benign users}.
The idea is inspired by \emph{matching networks} \cite{VinyalsBLKW16}, a popular meta-learning framework exploiting recent advances in attention mechanisms and external memories.

A matching network uses the output of an embedding model $f_\theta(x)$ to find similarities between input examples and reference examples from a \emph{support set}.
This non-parametric design, with external memories, allows matching networks to switch to a different classification task without supervised fine-tuning of $f_\theta$.
Specifically, given the trained embedding model $f_\theta(x)$ and a support set $\mathcal{S} = \{(x_i, y_i)\}_{i=1}^k$, class $\hat y = \arg\max_{i=1,\dots,k} P(y_i|\hat x,\mathcal S)$ is predicted where $P(y_i|\hat x,\mathcal S)$ estimates output probabilities for the input $\hat x$.
A common model is $\hat y = \sum_i a(\hat x,x_i) y_i$, a mixture of one-hot output vectors $y_i$ of the support set based on some \emph{attention mechanism} $a(\hat x, x_i)$ \cite{BahdanauCB14, LuongPM15, VaswaniSPUJGKP17, CollierB18, VinyalsBLKW16, SnellSZ17}.
For example, $a(\hat x, x_i)$ can be a softmax over the cosine distance $c(\cdot,\cdot)$ of the embeddings of $\hat x$ and $x_i$, i.e., $
  a(\hat x, x_i) = e^{c(f_\theta(\hat x), f_\theta(x_i))}/\big(\sum_{j=1}^k e^{c(f_\theta(\hat x), f_\theta(x_j))}\big)
$.
%
%During training, the parameters $\theta$ of the embedding model $f_\theta$ (e.g., a neural network without final classification layer) are updated to minimize the distance of the embeddings of training examples from those of examples with the same class in the support set.
%
%With external memories, matching networks can switch to a different classification task simply by using the embeddings of examples in the new support set, without supervised fine-tuning of $f_\theta$.
%
%An important characteristic of matching networks is that the classification of an input is not entirely determined by the parametric model $f_\theta$, but also by the matching procedure comparing embeddings.
%
We adopt a variant where (1)~the output components of the embedding model $f_\theta$ are multiplied by gate variables $0 \leq \alpha_{l,j} \leq 1$, and (2)~cosine distances to reference examples of each class are multiplied by a scaling factor $\beta_l$ \cite{Chen19}. Our attention mechanism is thus a softmax over embedding distances $c(\alpha_l \odot f_\theta(\hat x), f_\theta(x))\beta_l$.

Our defense mechanism requires fine-tuning to train learnable parameters $\alpha_l$ and $\beta_l$.
Before fine-tuning the adapted matching network, we apply a random Glorot initialization ${\theta_{g}}$
 as ${\theta'} = {\delta}{\theta} + (1-{\delta}){\theta_{g}}$
to reduce the influence of the poisoned model; then we train $\alpha_l$ and $\beta_l$ for a few iterations (with fixed $\theta'$), and finally train $\theta'$, $\alpha_l$ and $\beta_l$ jointly (training is performed as in \cite[Sec.~4.1]{VinyalsBLKW16}).
We use $\delta=0.3$ and the same learning rate $\eta = 0.001$ of backdoor experiments in \cref{sec:bd_result}.
Note that this \emph{fine-tuning is not necessary for matching networks but provides a defense against backdoor attacks}, as it allows our method to remove anomalies introduced in the embedding model $f_\theta(x)$ by the attacker.

\noindent\textbf{Experiment 3.}\hspace{.5em}
%Using the same settings as in Experiment 2, we perform ablation analysis for Glorot initialization and matching networks. Due to lack of space, here we report results for $\delta=0.3$ with the adapted matching networks as in \cref{fig:005,fig:005m}.
%(columns correspond to clients and rows to use cases of correctly labeled backdoor examples; fine-tuning starts at epoch 2).
% Chien-Lun: This is duplicated; we have mentioned this in Experiment 2, so I remove this to save some space.
% Chien-Lun: we also forgot to mention the setting of Experiment 3, so I revised it a little bit.
%
Results are reported in \cref{fig:005,fig:005m} (fine-tuning starts in Round 2, using the same parameters as in Experiment~2).
\emph{The proposed defense mechanism can successfully remove backdoor attacks:} when backdoor classes are not present in meta-testing (\cref{fig:005_1,fig:005_2,fig:005m_1,fig:005m_2}), attack accuracy drops to $\approx 20\%$ (comparable to random assignment to one of the 5 classes) in a few epochs;
when backdoor classes are present in meta-testing (\cref{fig:005_3,fig:005m_3}), attack accuracy significantly drops to $\approx 0\%$ (Omniglot) and $\approx 10\%$ (mini-ImageNet) in a few epochs of fine-tuning.
Notably, meta-testing accuracy for Omniglot (\cref{fig:005}) is always above $96\%$ after $50$ iterations;
in contrast, meta-testing accuracy for mini-ImageNet (\cref{fig:005m}) is $\approx 35\%$, lower than in \cref{fig:002m}.
This suggests a limitation of matching networks; other variants may overcome this limitation.
%, or non-parametric classification mechanisms such as Prototypical Networks \cite{SnellSZ17} and Relation Networks \cite{SungYZXTH18}, could overcome this limitation.
%
An ablation study of our defense mechanism (see Appendix) highlighted the importance of all of its elements (attention model, noisy meta-model initialization, fine-tuning procedure).

%% file: tex/conclusions.tex
\section{Conclusions}
\label{sec:conclusions}

We showed that one-shot poisoning backdoor attacks can be very successful in federated \emph{meta-learning}, even on backdoor class examples not used by the attacker and after additional meta-learning or long fine-tuning by benign users.
We presented a defense mechanism based on matching networks, compatible with secure update aggregation at the server and effective in eliminating the attack, but with some main-task accuracy reduction.
Our future efforts will focus on this limitation.

%% file: tex/appendix.tex
\section*{Appendix}
\renewcommand{\thesection}{\Alph{section}} 
\renewcommand{\thefigure}{A\arabic{figure}}
\setcounter{section}{0}
\setcounter{figure}{0}

In the appendix, we provide additional experimental details and results, including:
\begin{itemize} 
    \item Benign fine-tuning of matching networks after attacks with different values of $\delta$.
    \item Benign (supervised) fine-tuning after attacks with $\delta=0.3$.
    \item Extra benign \emph{local meta-training} before benign fine-tuning (supervised/matching networks) after attacks for $\delta=0.3$.
    \item Experimental setup and run time.
\end{itemize}

\section{Additional Experimental Results}
\subsection{Benign fine-tuning of matching networks with different $\delta$}

\begin{figure}[th!]
     \centering
     \begin{minipage}[t]{.495\textwidth}
     \centering
     \begin{subfigure}[hbt!]{\textwidth}
         \centering
         \includegraphics[width=\textwidth]{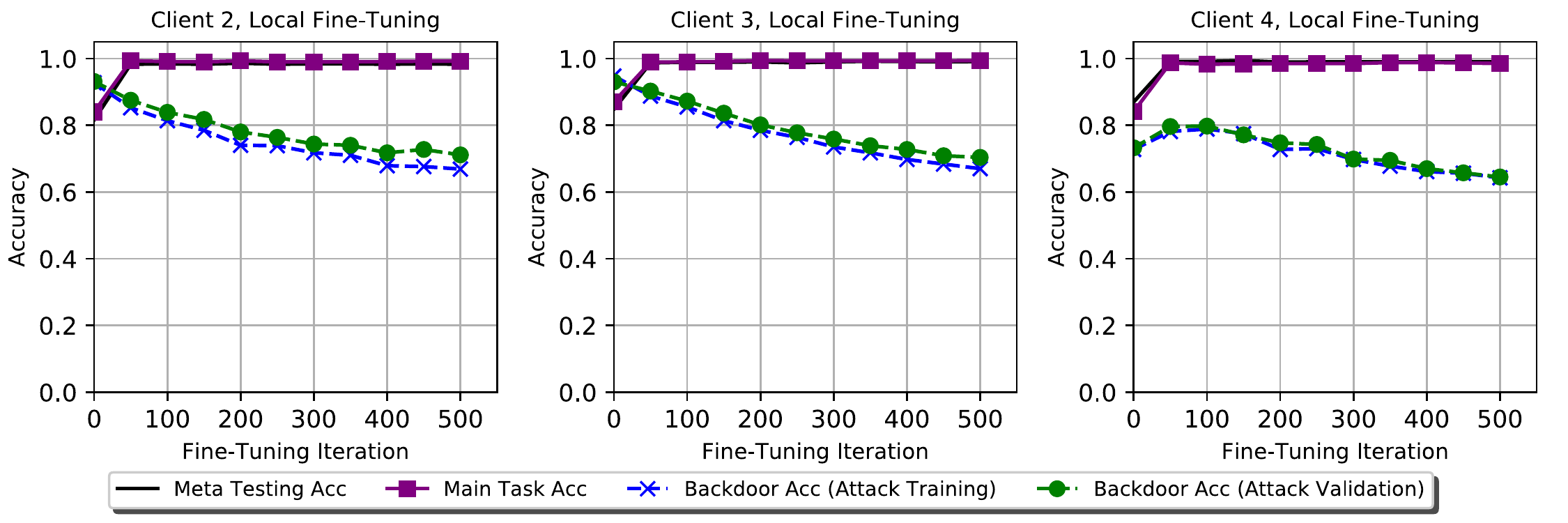}
         \vspace*{-6mm}
         \caption{Backdoor examples not used by benign users}
         \label{fig_ap:005_1}
     \end{subfigure}
     \hfill
     \vspace*{1mm}
     \begin{subfigure}[hbt!]{\textwidth}
         \centering
         \includegraphics[width=\textwidth]{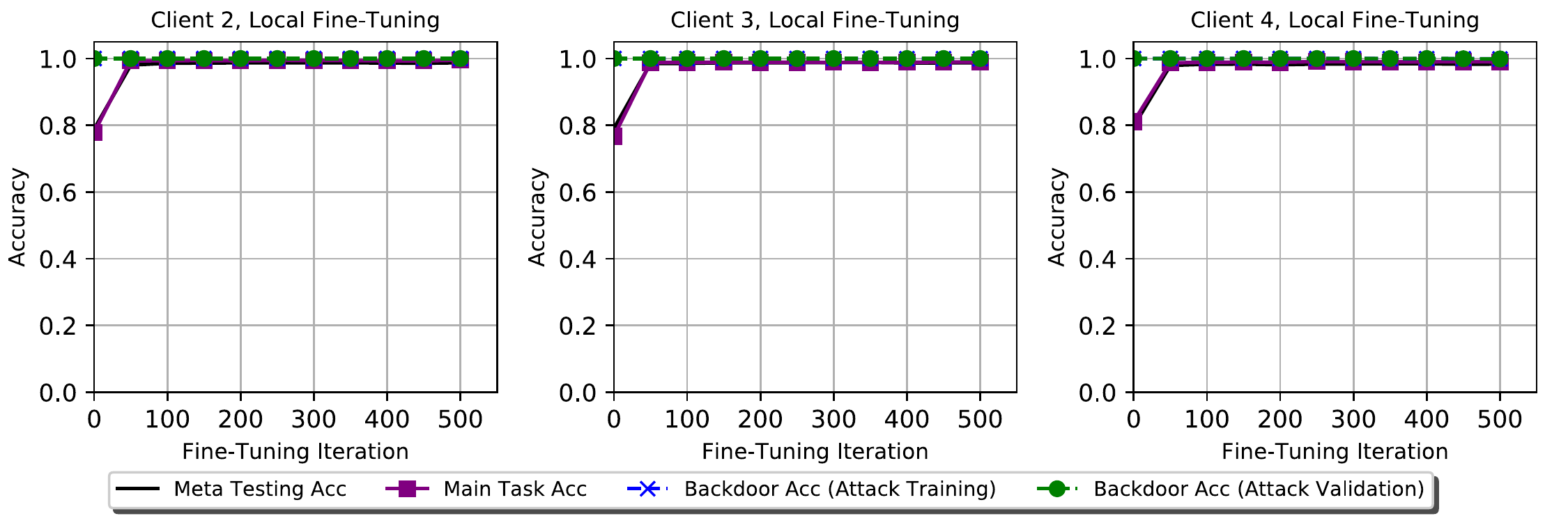}
         \vspace*{-6mm}
         \caption{Backdoor classes used in benign pre-training}
         \label{fig_ap:005_2}
     \end{subfigure}
     \hfill
     \vspace*{1mm}
     \begin{subfigure}[hbt!]{\textwidth}
         \centering
         \includegraphics[width=\textwidth]{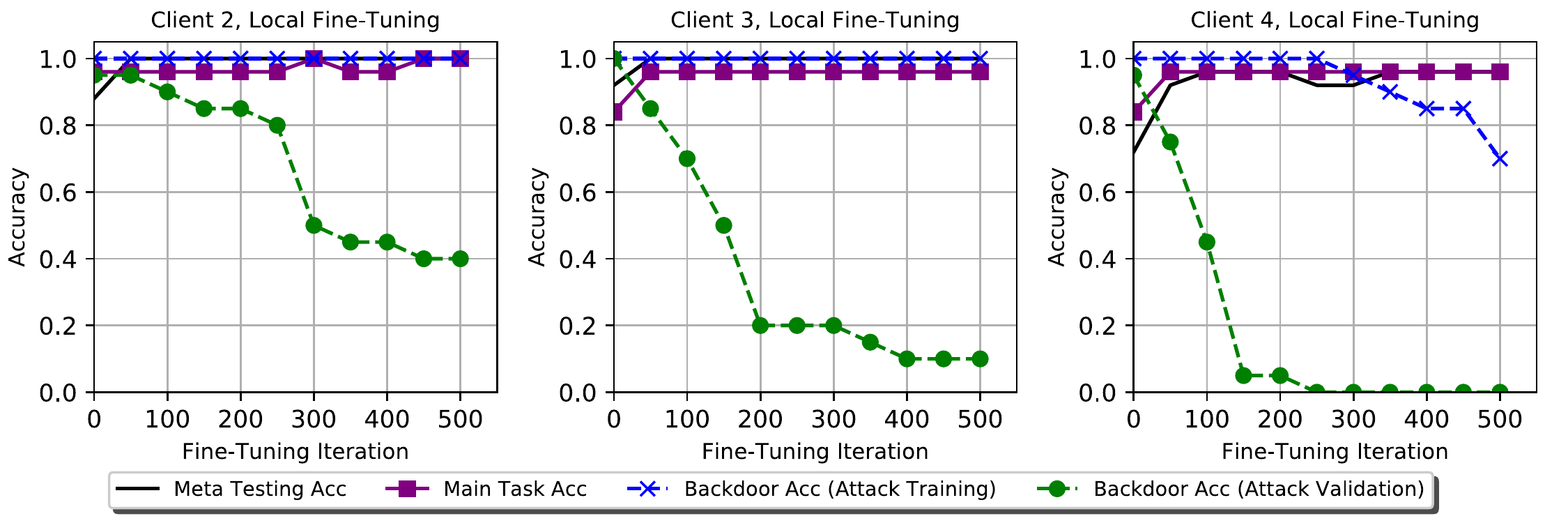}
         \vspace*{-6mm}
         \caption{Backdoor classes used also in benign fine-tuning}
         \label{fig_ap:005_3}
     \end{subfigure}
     % \vspace*{-1mm}
     \caption{Benign fine-tuning of matching networks ($\eta = 0.001, \delta=0.6$) after attacks on Omniglot}
     \vspace*{-3mm}
     \label{fig_ap:005}
     \end{minipage}
     \hfill
     \centering
     \begin{minipage}[t]{.495\textwidth}
     \centering
     \begin{subfigure}[hbt!]{\textwidth}
         \centering
         \includegraphics[width=\textwidth]{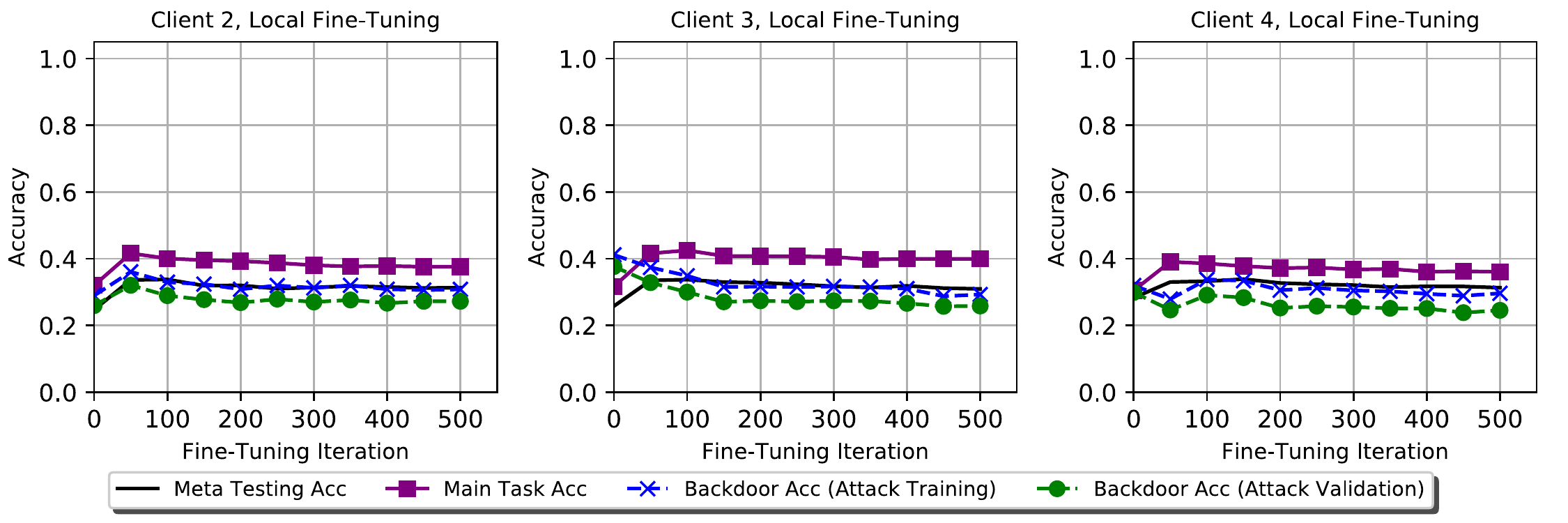}
         \vspace*{-6mm}
         \caption{Backdoor examples not used by benign users}
         \label{fig_ap:005m_1}
     \end{subfigure}
     \hfill
     \vspace*{1mm}
     \begin{subfigure}[hbt!]{\textwidth}
         \centering
         \includegraphics[width=\textwidth]{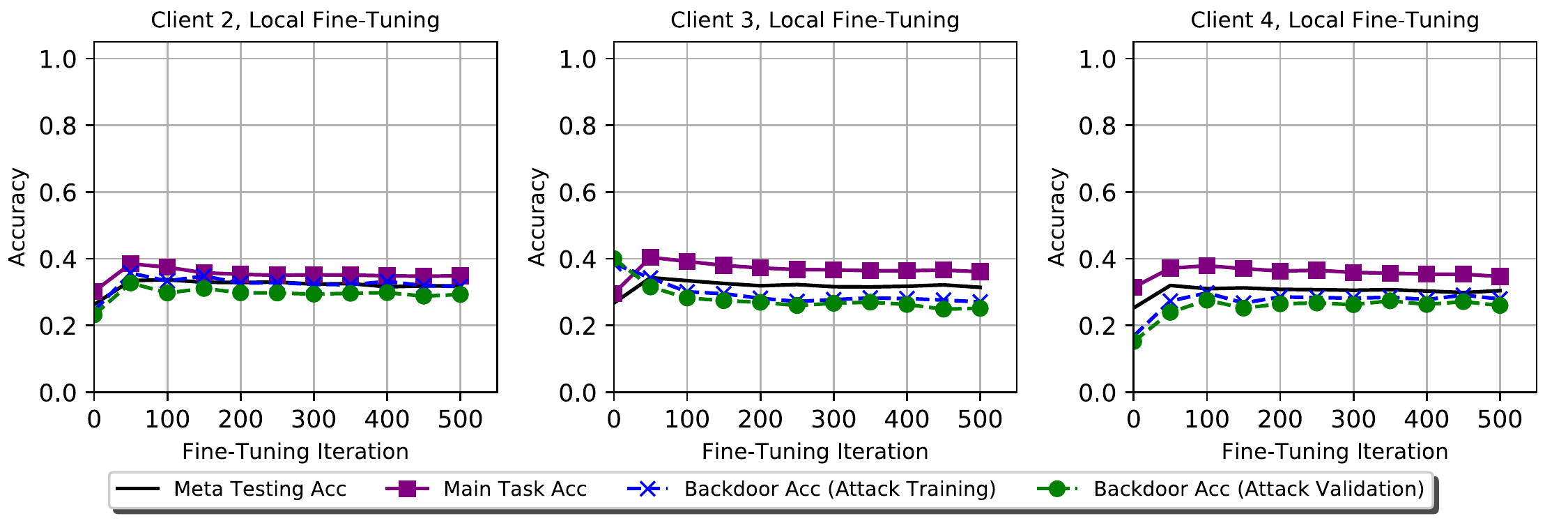}
         \vspace*{-6mm}
         \caption{Backdoor classes used in benign pre-training}
         \label{fig_ap:005m_2}
     \end{subfigure}
     \hfill
     \vspace*{1mm}
     \begin{subfigure}[hbt!]{\textwidth}
         \centering
         \includegraphics[width=\textwidth]{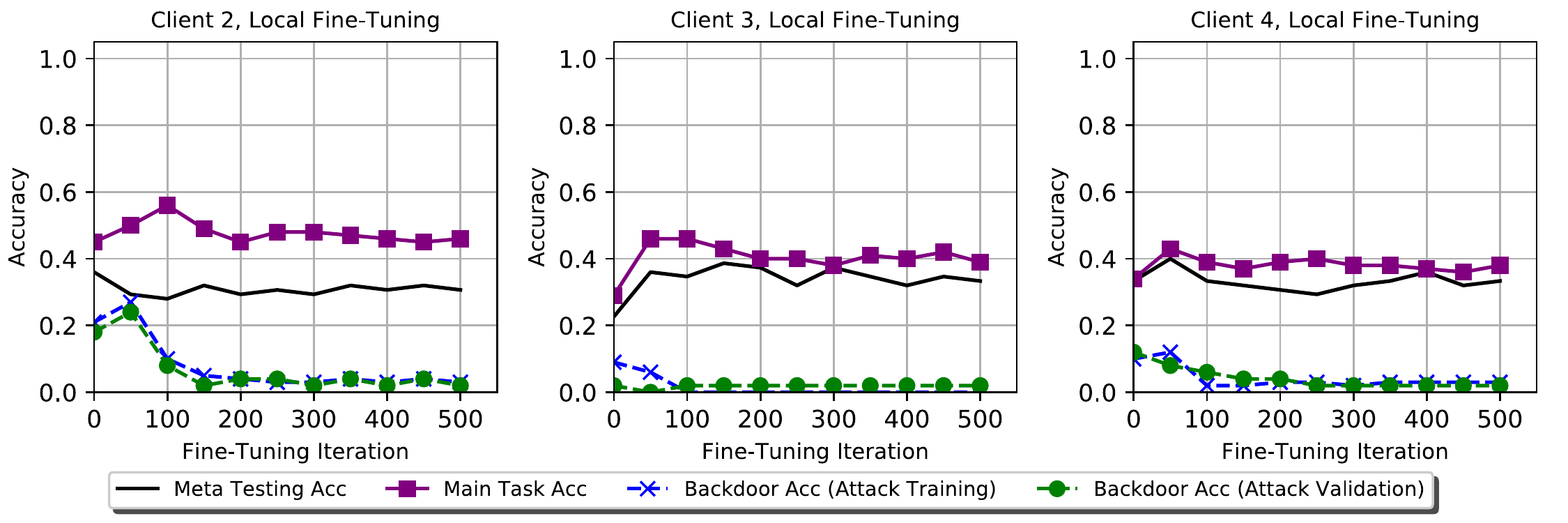}
         \vspace*{-6mm}
         \caption{Backdoor classes used also in benign fine-tuning}
         \label{fig_ap:005m_3}
     \end{subfigure}
     % \vspace*{-1mm}
     \caption{Benign fine-tuning of matching networks ($\eta = 0.001, \delta=0.6$) after attacks on mini-ImageNet}
     \vspace*{-3mm}
     \label{fig_ap:005m}
     \end{minipage}
     % \vspace*{3mm}
\end{figure}

\begin{figure}[hbt!]
     \centering
     \begin{minipage}[t]{.495\textwidth}
     \centering
     \begin{subfigure}[hbt!]{\textwidth}
         \centering
         \includegraphics[width=\textwidth]{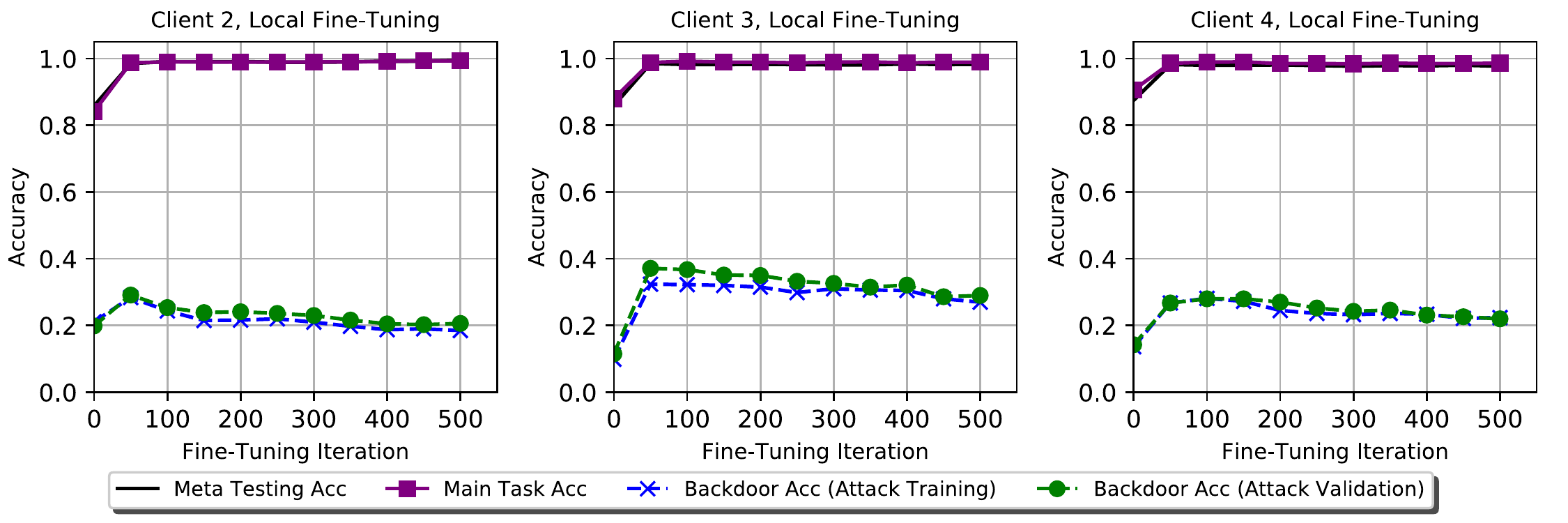}
         \vspace*{-6mm}
         \caption{Backdoor examples not used by benign users}
         \label{fig_ap:006_1}
     \end{subfigure}
     \hfill
     \vspace*{1mm}
     \begin{subfigure}[hbt!]{\textwidth}
         \centering
         \includegraphics[width=\textwidth]{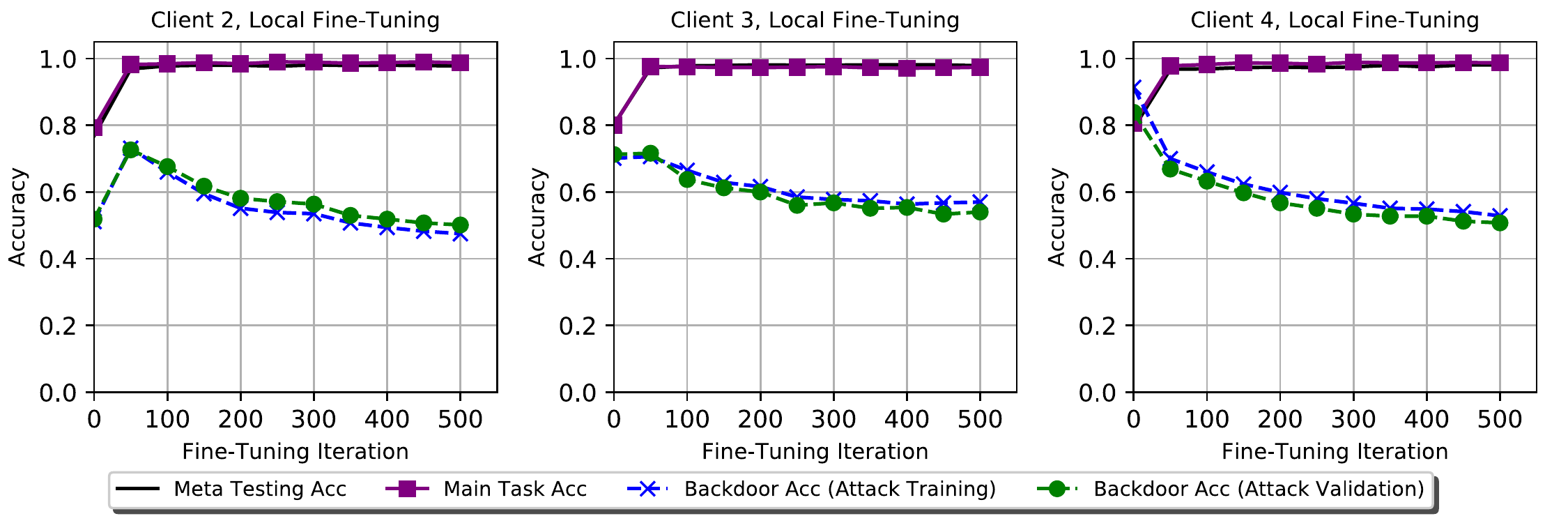}
         \vspace*{-6mm}
         \caption{Backdoor classes used in benign pre-training}
         \label{fig_ap:006_2}
     \end{subfigure}
     \hfill
     \vspace*{1mm}
     \begin{subfigure}[hbt!]{\textwidth}
         \centering
         \includegraphics[width=\textwidth]{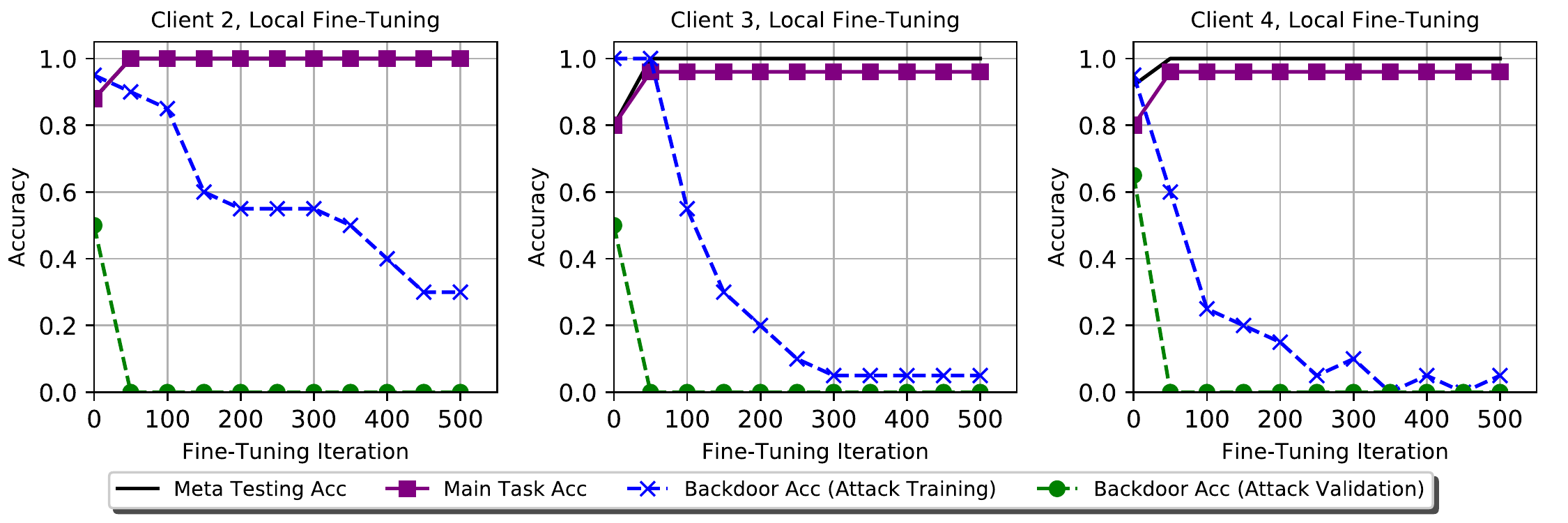}
         \vspace*{-6mm}
         \caption{Backdoor classes used also in benign fine-tuning}
         \label{fig_ap:006_3}
     \end{subfigure}
     % \vspace*{-1mm}
     \caption{Benign fine-tuning of matching networks ($\eta = 0.001, \delta=0.4$) after attacks on Omniglot}
     \vspace*{-3mm}
     \label{fig_ap:006}
     \end{minipage}
          \hfill
     \centering
     \begin{minipage}[t]{.495\textwidth}
     \centering
     \begin{subfigure}[hbt!]{\textwidth}
         \centering
         \includegraphics[width=\textwidth]{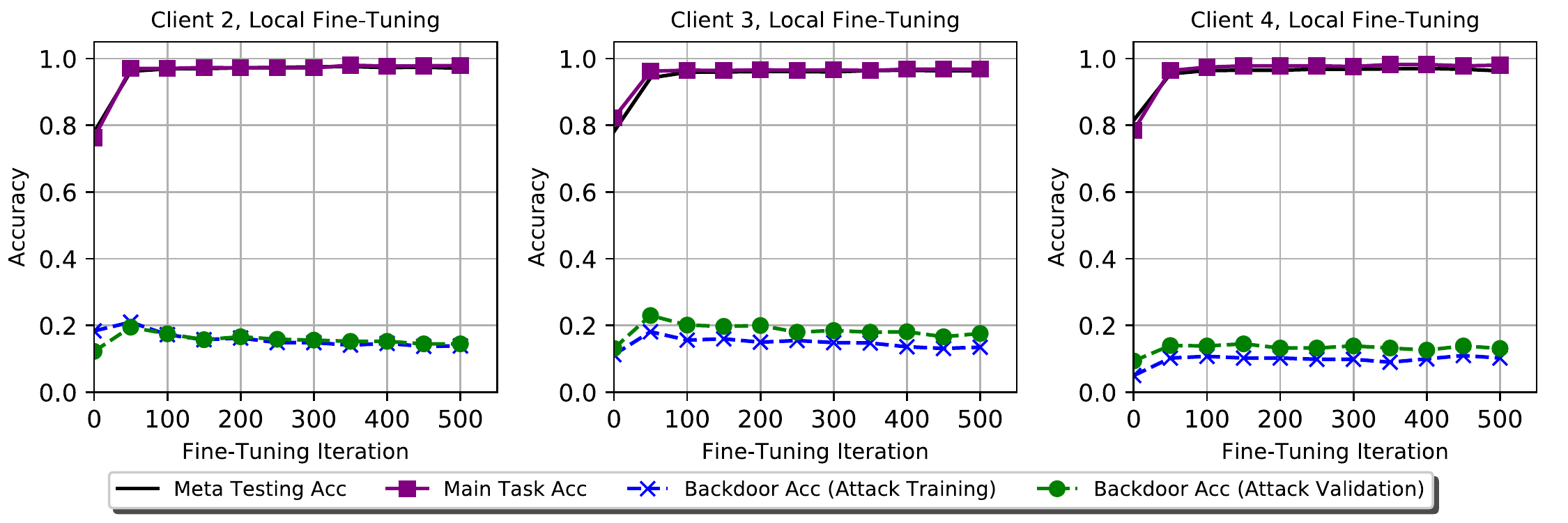}
         \vspace*{-6mm}
         \caption{Backdoor examples not used by benign users}
         \label{fig_ap:007_1}
     \end{subfigure}
     \hfill
     \vspace*{1mm}
     \begin{subfigure}[hbt!]{\textwidth}
         \centering
         \includegraphics[width=\textwidth]{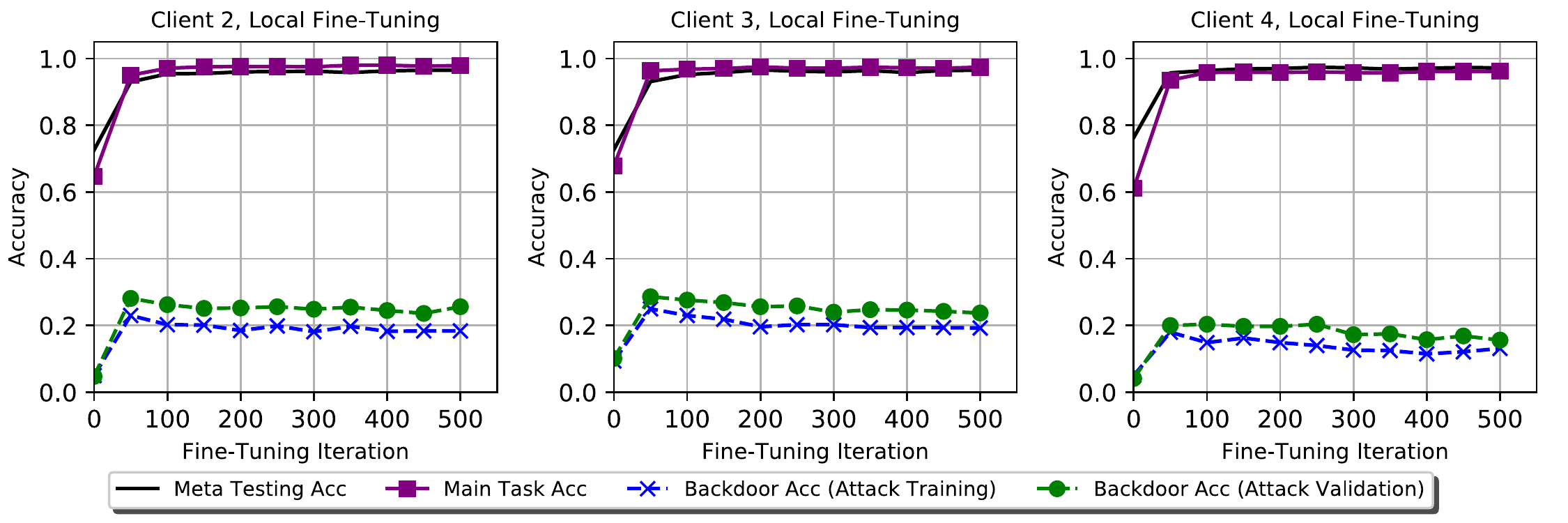}
         \vspace*{-6mm}
         \caption{Backdoor classes used in benign pre-training}
         \label{fig_ap:007_2}
     \end{subfigure}
     \hfill
     \vspace*{1mm}
     \begin{subfigure}[hbt!]{\textwidth}
         \centering
         \includegraphics[width=\textwidth]{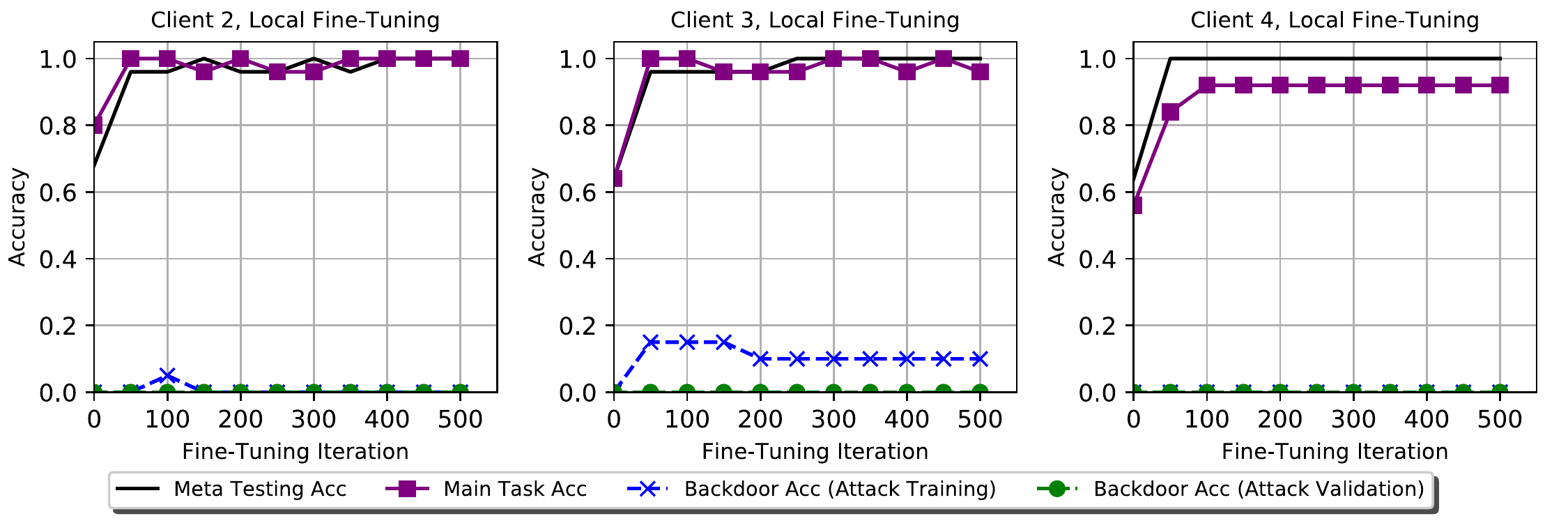}
         \vspace*{-6mm}
         \caption{Backdoor classes used also in benign fine-tuning}
         \label{fig_ap:007_3}
     \end{subfigure}
     % \vspace*{-1mm}
     \caption{Benign fine-tuning of matching networks ($\eta = 0.001, \delta=0.2$) after attacks on Omniglot}
     \vspace*{-3mm}
     \label{fig_ap:007}
     \end{minipage}
     \vspace*{1mm}
\end{figure}

In the paper, we demonstrate results of benign fine-tuning of matching networks for $\delta=0.3$.

\begin{figure}[t!]
     \centering
     % \begin{minipage}[t]{.495\textwidth}
     \centering
     \begin{subfigure}[hbt!]{.48\textwidth}
         \centering
         \includegraphics[width=\textwidth]{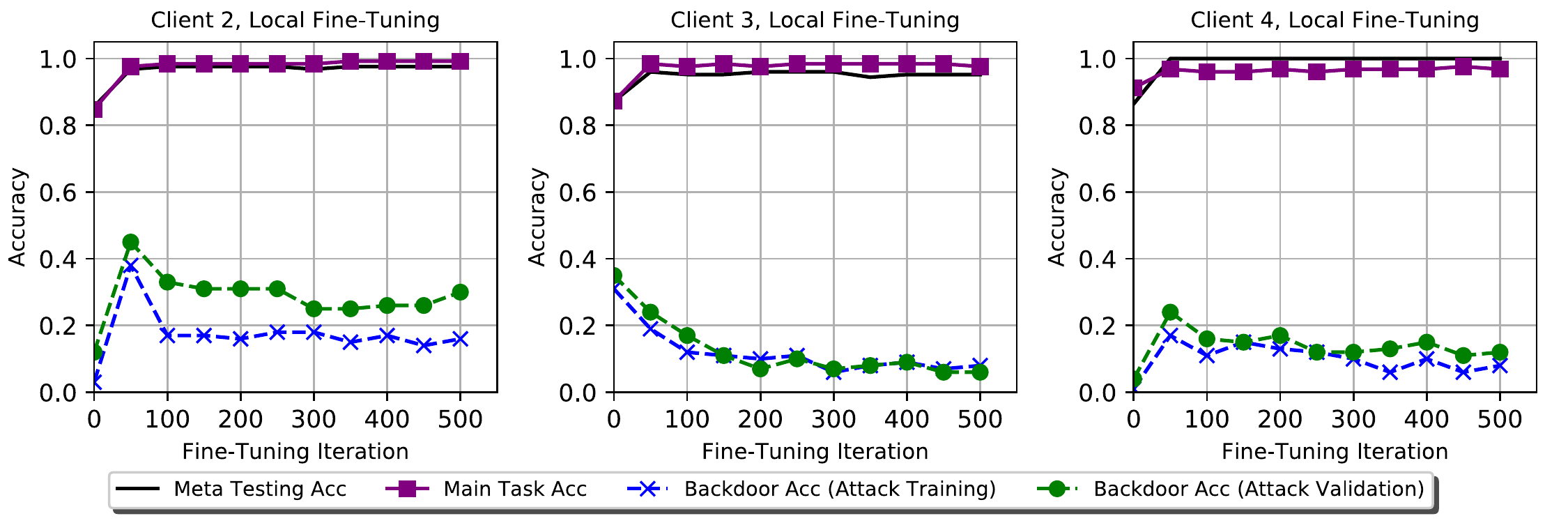}
         \vspace*{-6mm}
         \caption{Backdoor examples not used by benign users}
         \label{fig_ap:008_extra_1}
     \end{subfigure}
     \\
     \vspace*{1mm}
     \begin{subfigure}[hbt!]{.48\textwidth}
         \centering
         \includegraphics[width=\textwidth]{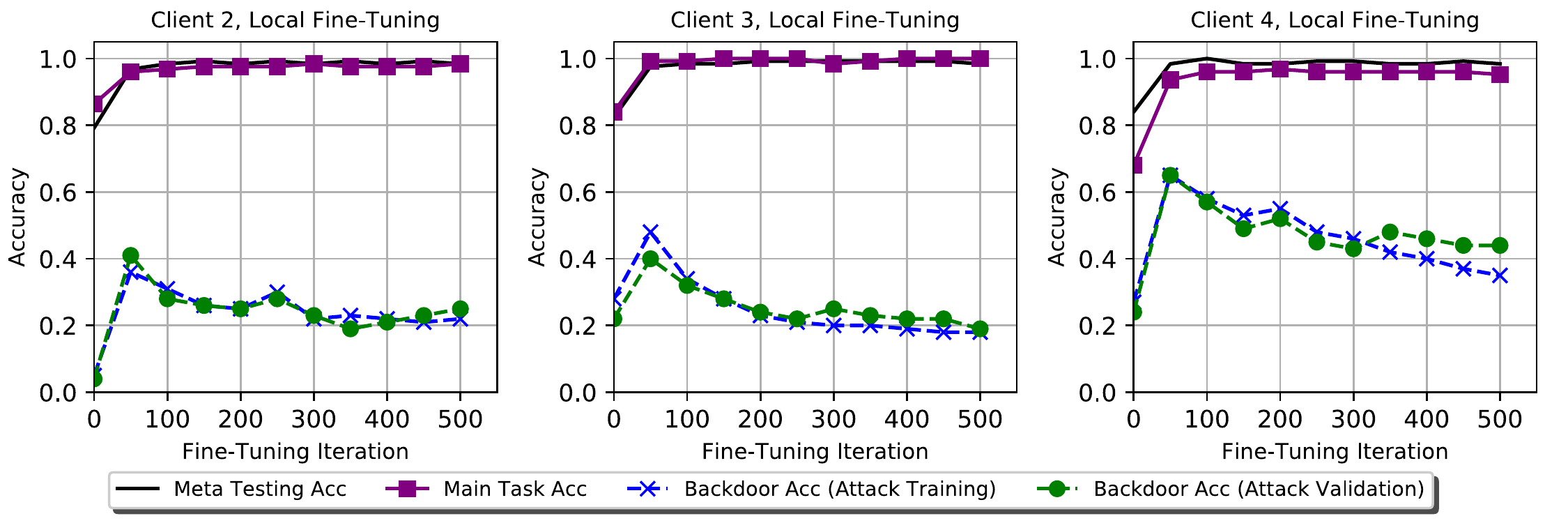}
         \vspace*{-6mm}
         \caption{Backdoor classes used in benign pre-training}
         \label{fig_ap:008_extra_2}
     \end{subfigure}
     \\
     \vspace*{1mm}
     \begin{subfigure}[hbt!]{.48\textwidth}
         \centering
         \includegraphics[width=\textwidth]{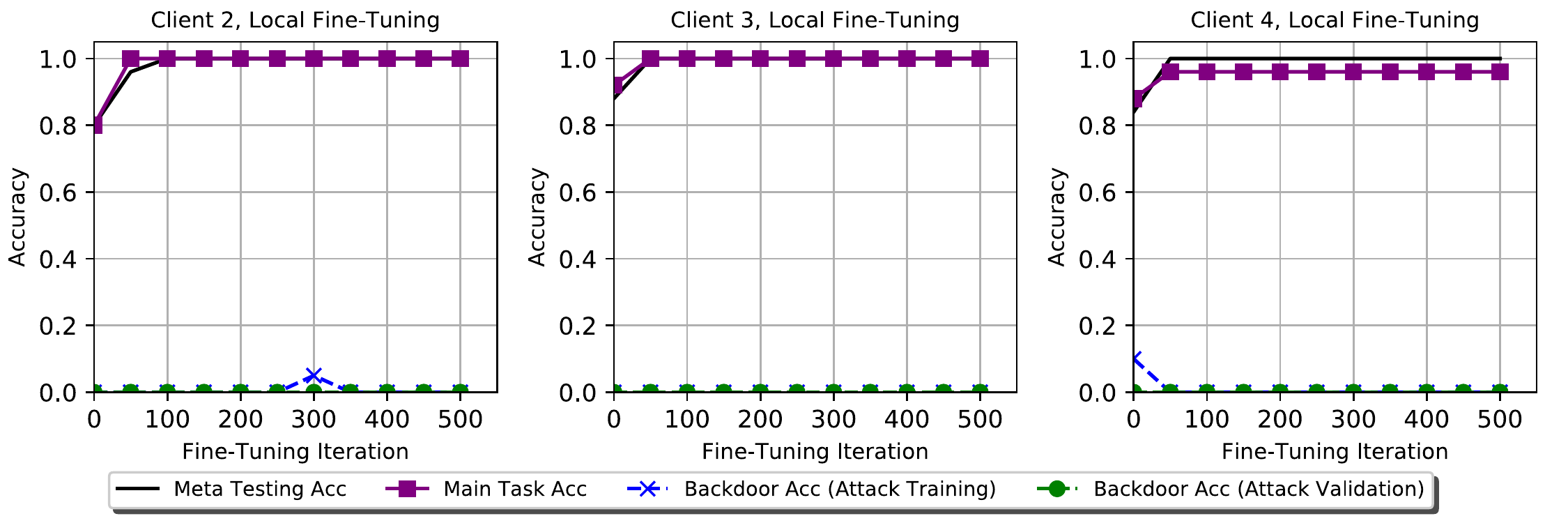}
         \vspace*{-6mm}
         \caption{Backdoor classes used also in benign fine-tuning}
         \label{fig_ap:008_extra_3}
     \end{subfigure}
     % \vspace*{-1mm}
     \caption{Benign fine-tuning of matching networks ($\eta = 0.001, \delta=0.3$) after attacks on Omniglot (different set of $40$ episodes)}
     \vspace*{-3mm}
     \label{fig_ap:008_extra}
     % \end{minipage}
     % \vspace*{3mm}
\end{figure}

In \cref{fig_ap:005} and \cref{fig_ap:005m}, we report results of matching networks fine-tuning with $\delta = 0.6$ for Omniglot and mini-ImageNet, respectively (larger $\delta$ implies less randomness). 
Effects of backdoor attacks can still be removed effectively in mini-ImageNet (\cref{fig_ap:005m}): main-task accuracy and meta-testing accuracy are similar to the case of $\delta=0.3$ (\cref{fig:005m}, or \cref{fig_ap:008_extra} for a different set of $40$ episodes). However, effects of backdoor attacks cannot be removed in Omniglot with $\delta=0.6$; results for Omniglot with smaller values of $\delta$ ($0.4$ and $0.2$) are also reported in \cref{fig_ap:006} and \cref{fig_ap:007}, respectively, for reference.

As expected, introducing more randomness (appropriately) can remove backdoor effects more effectively; however, introducing too much randomness can damage both main task accuracy and meta-testing accuracy due to the dominance of noise.
% while inadequately large randomness can damage both main task accuracy and meta-testing accuracy due to the dominance of noise. {\bf from LG: what do you mean "inadequately large randomness"? inadequetely and large seem to be opposites?} 
As shown in \cref{fig_ap:007_1,fig_ap:007_2}, for $\delta=0.2$, main-task accuracy and meta-testing accuracy are $3\%$ lower than in \cref{fig_ap:006_1,fig_ap:006_2} ($\delta=0.4$). In light of this, finding an appropriate value of $\delta$ (or, equivalently, ratio between the norms of a trained model and an initialization) is an interesting problem and part of our future efforts.
% {\bf from LG: so, is finding appropriate value of delta for a particular dataset part of future work?}

\subsection{Benign (supervised) fine-tuning after attacks for $\delta=0.3$}

\begin{figure}[t!]
     \centering
     \begin{minipage}[t]{.495\textwidth}
     \centering
     \begin{subfigure}[hbt!]{\textwidth}
         \centering
         \includegraphics[width=\textwidth]{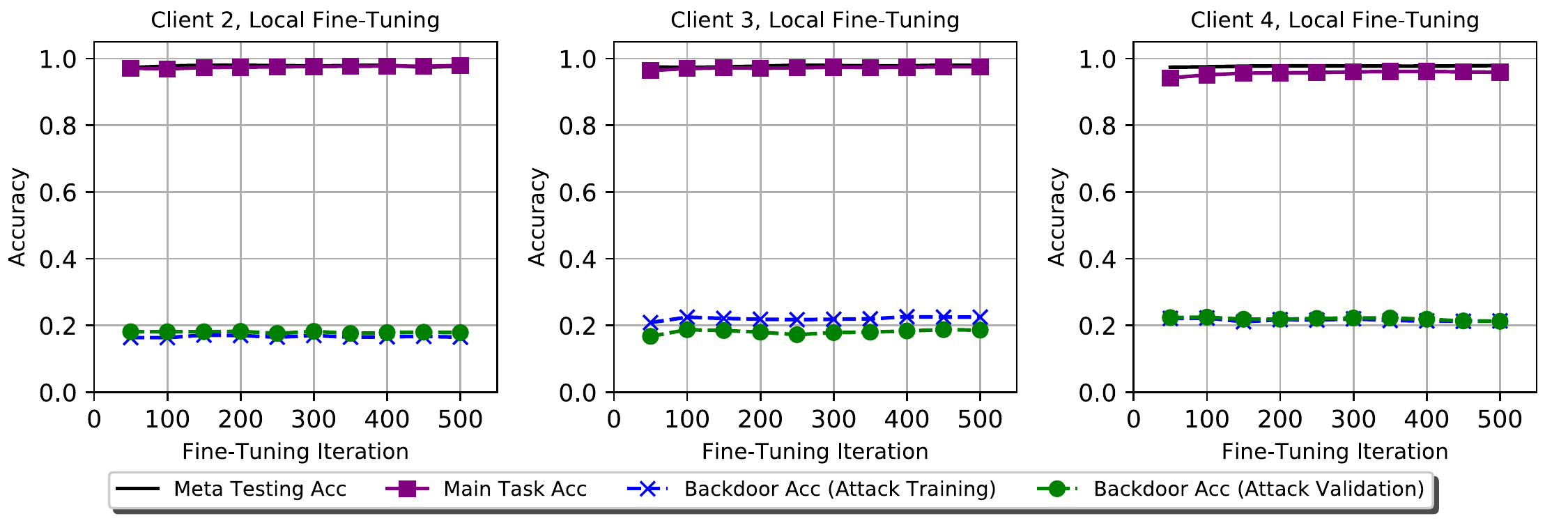}
         \vspace*{-6mm}
         \caption{Backdoor examples not used by benign users}
         \label{fig_ap:008_1}
     \end{subfigure}
     \hfill
     \vspace*{1mm}
     \begin{subfigure}[hbt!]{\textwidth}
         \centering
         \includegraphics[width=\textwidth]{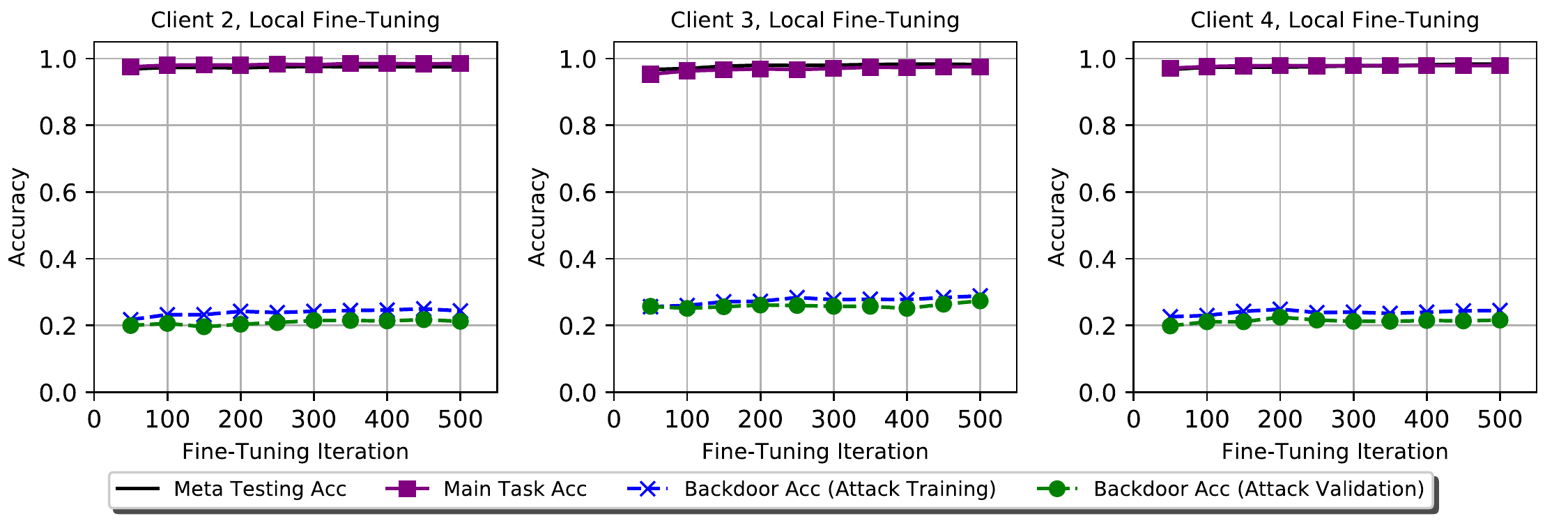}
         \vspace*{-6mm}
         \caption{Backdoor classes used in benign pre-training}
         \label{fig_ap:008_2}
     \end{subfigure}
     \hfill
     \vspace*{1mm}
     \begin{subfigure}[hbt!]{\textwidth}
         \centering
         \includegraphics[width=\textwidth]{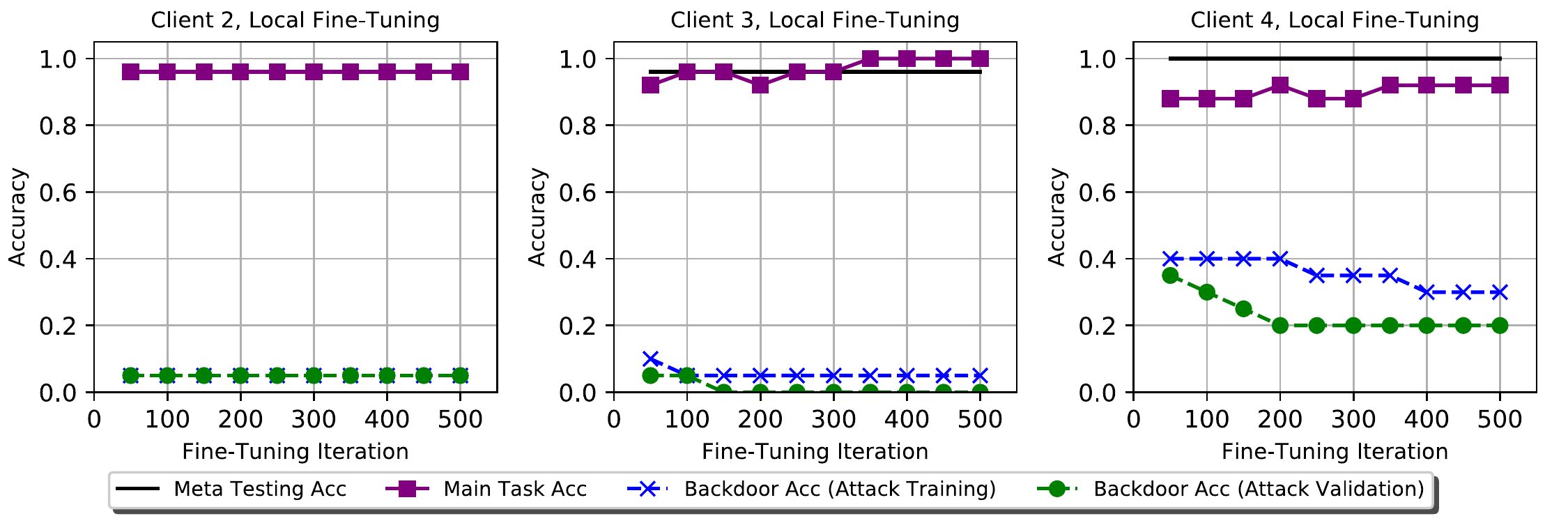}
         \vspace*{-6mm}
         \caption{Backdoor classes used also in benign fine-tuning}
         \label{fig_ap:008_3}
     \end{subfigure}
     % \vspace*{-1mm}
     \caption{Benign fine-tuning ($\eta = 0.001, \delta=0.3$) after attacks on Omniglot}
     \vspace*{-3mm}
     \label{fig_ap:008}
     \end{minipage}
     \hfill
     \centering
     \begin{minipage}[t]{.495\textwidth}
     \centering
     \begin{subfigure}[hbt!]{\textwidth}
         \centering
         \includegraphics[width=\textwidth]{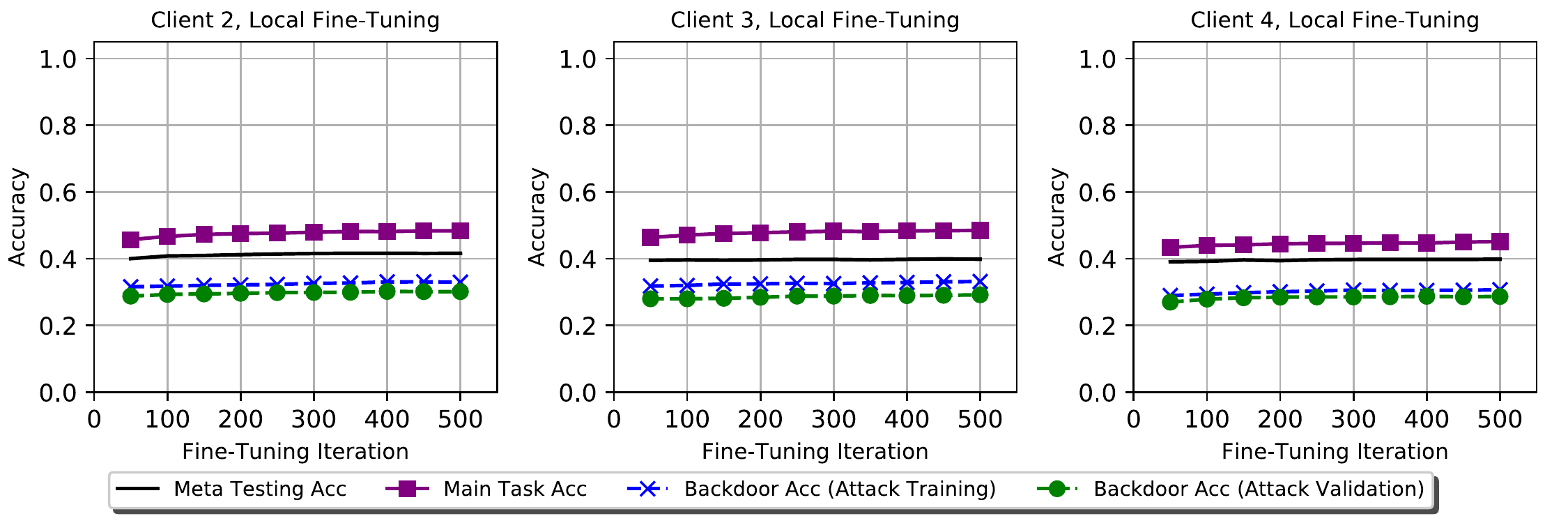}
         \vspace*{-6mm}
         \caption{Backdoor examples not used by benign users}
         \label{fig_ap:008m_1}
     \end{subfigure}
     \hfill
     \vspace*{1mm}
     \begin{subfigure}[hbt!]{\textwidth}
         \centering
         \includegraphics[width=\textwidth]{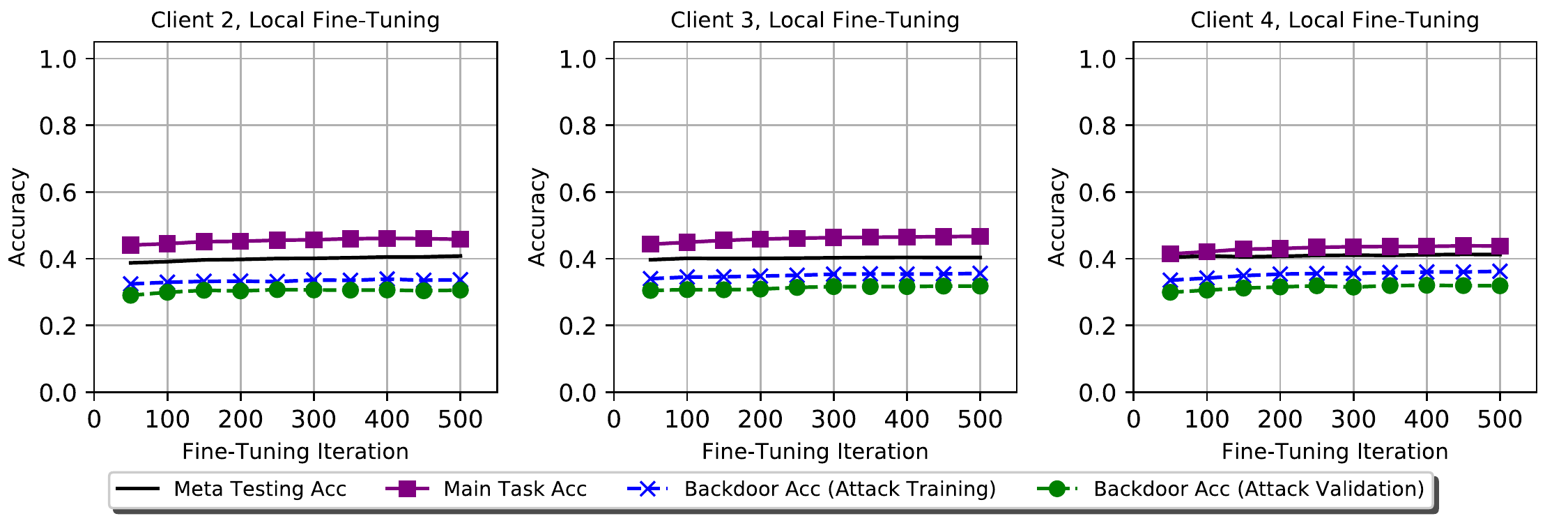}
         \vspace*{-6mm}
         \caption{Backdoor classes used in benign pre-training}
         \label{fig_ap:008m_2}
     \end{subfigure}
     \hfill
     \vspace*{1mm}
     \begin{subfigure}[hbt!]{\textwidth}
         \centering
         \includegraphics[width=\textwidth]{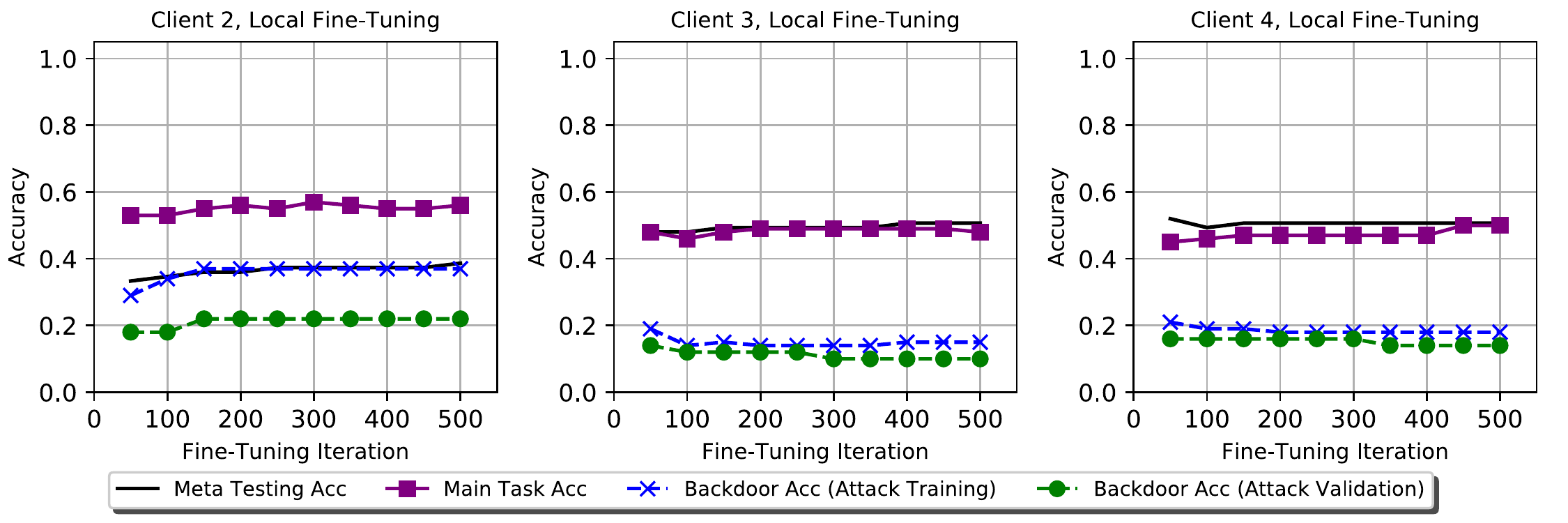}
         \vspace*{-6mm}
         \caption{Backdoor classes used also in benign fine-tuning}
         \label{fig_ap:008m_3}
     \end{subfigure}
     % \vspace*{-1mm}
     \caption{Benign fine-tuning ($\eta = 0.001, \delta=0.3$) after attacks on mini-ImageNet}
     \vspace*{-3mm}
     \label{fig_ap:008m}
     \end{minipage}
\end{figure}

In the paper, we demonstrate results of benign (supervised) fine-tuning after attacks (\cref{fig:002,fig:002m}, for Omniglot and mini-ImageNet, respectively). We also demonstrate results of benign fine-tuning of matching networks after attacks (\cref{fig:005,fig:005m}), in which random initialization parameters are introduced ($\delta=0.3$). Here we demonstrate results of benign supervised fine-tuning after attacks \emph{when random initialization parameters are introduced} ($\delta=0.3$ as well). 

Results are reported in \cref{fig_ap:008} and \cref{fig_ap:008m}. For Omniglot (\cref{fig_ap:008}), supervised fine-tuning performs similarly to our proposed fine-tuning of matching networks (\cref{fig:005}) except for client 4 in case (c). For mini-ImageNet, attack accuracy can only be reduced to $30\%$ through supervised fine-tuning (\cref{fig_ap:008m_1,fig_ap:008m_2}), while the use of matching networks (\cref{fig:005m_1,fig:005m_2}) can reduce it to $20\%$. Similarly, for case (c), attack accuracy can only be reduced to $20\%$ through supervised fine-tuning (\cref{fig_ap:008m_3}), while the use of matching networks (\cref{fig:005m_3}) can reduce it to $10\%$ (the initial attack accuracy before the backdoor attack, due to misclassification). 
This shows the effectiveness of our proposed fine-tuning of matching networks on removing effects of backdoor attacks, particularly when benign examples in backdoor classes are available during fine-tuning.
% {\bf from LG: so, what conclusions do we make from this? Our approach is still useful in some cases?}

\subsection{Extra benign local meta-training before benign fine-tuning (supervised/matching networks) after attacks for $\delta=0.3$}

In the previous sections, we have shown that proper randomness allows fine-tuning of matching networks to remove the effects of backdoor attacks. In this section, we evaluate whether running extra \emph{local meta-training} based on local user data before fine-tuning can improve main-task accuracy and meta-testing accuracy in mini-ImageNet (where it is lower).

\begin{figure}[t!]
     \centering
     \begin{minipage}[t]{.495\textwidth}
     \centering
     \begin{subfigure}[hbt!]{\textwidth}
         \centering
         \includegraphics[width=\textwidth]{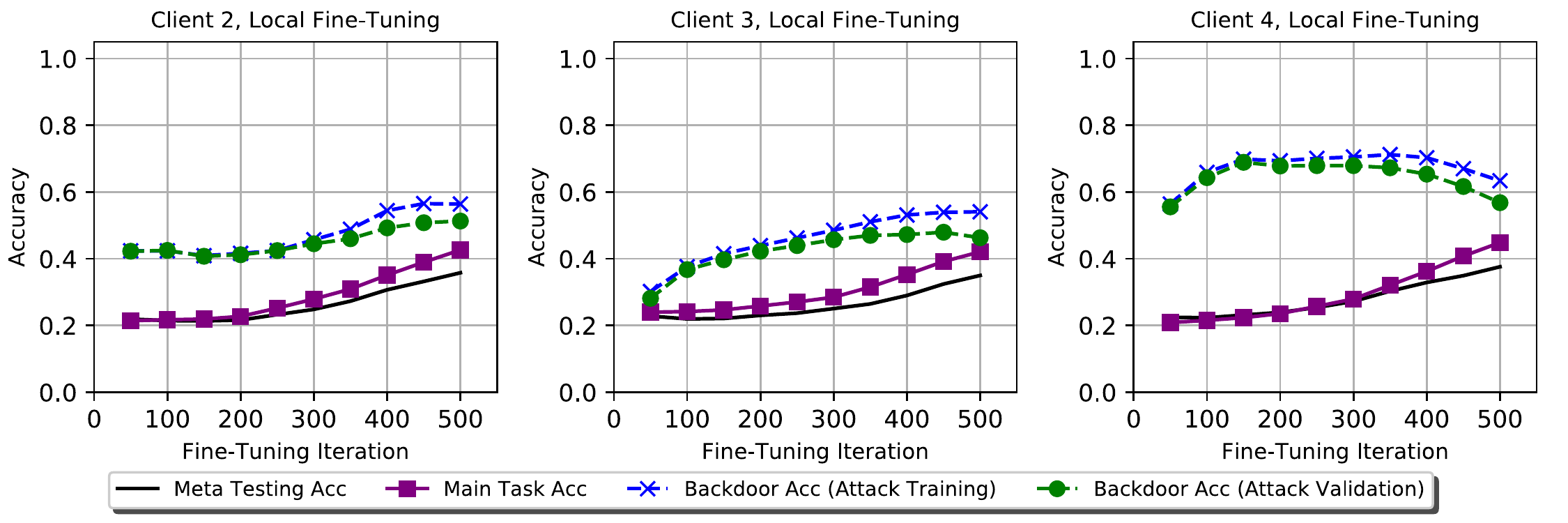}
         \vspace*{-6mm}
         \caption{Backdoor examples not used by benign users}
         \label{fig_ap:009_1}
     \end{subfigure}
     \hfill
     \vspace*{1mm}
     \begin{subfigure}[hbt!]{\textwidth}
         \centering
         \includegraphics[width=\textwidth]{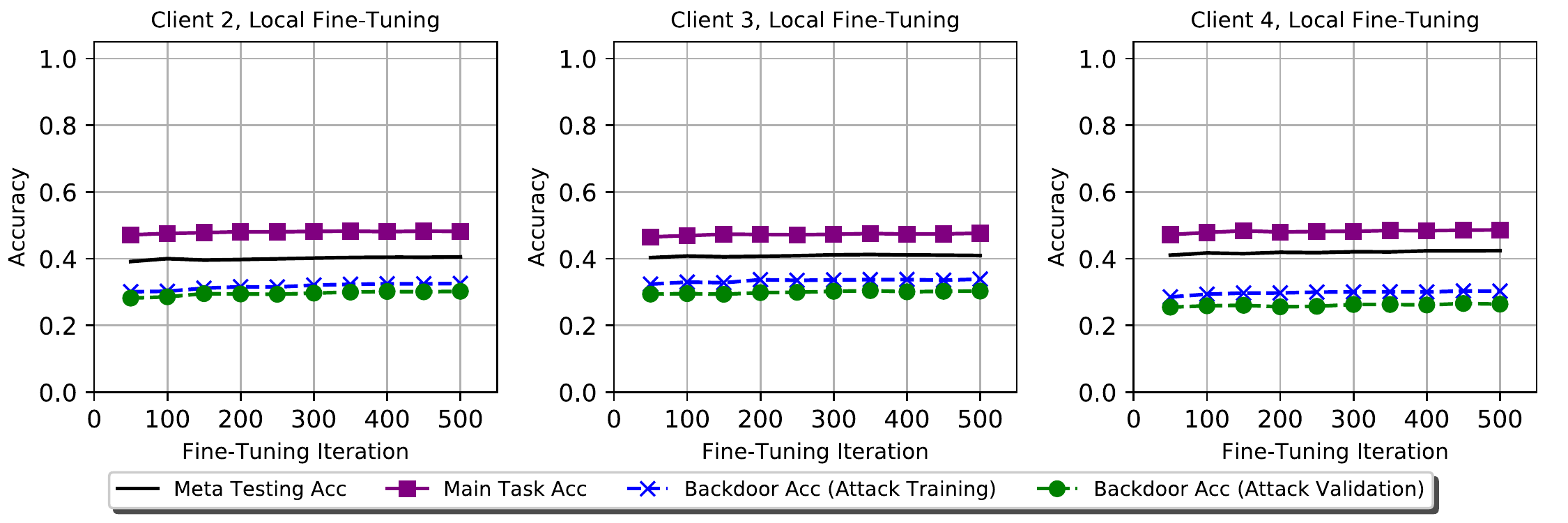}
         \vspace*{-6mm}
         \caption{Backdoor classes used in benign pre-training}
         \label{fig_ap:009_2}
     \end{subfigure}
     \hfill
     \vspace*{1mm}
     \begin{subfigure}[hbt!]{\textwidth}
         \centering
         \includegraphics[width=\textwidth]{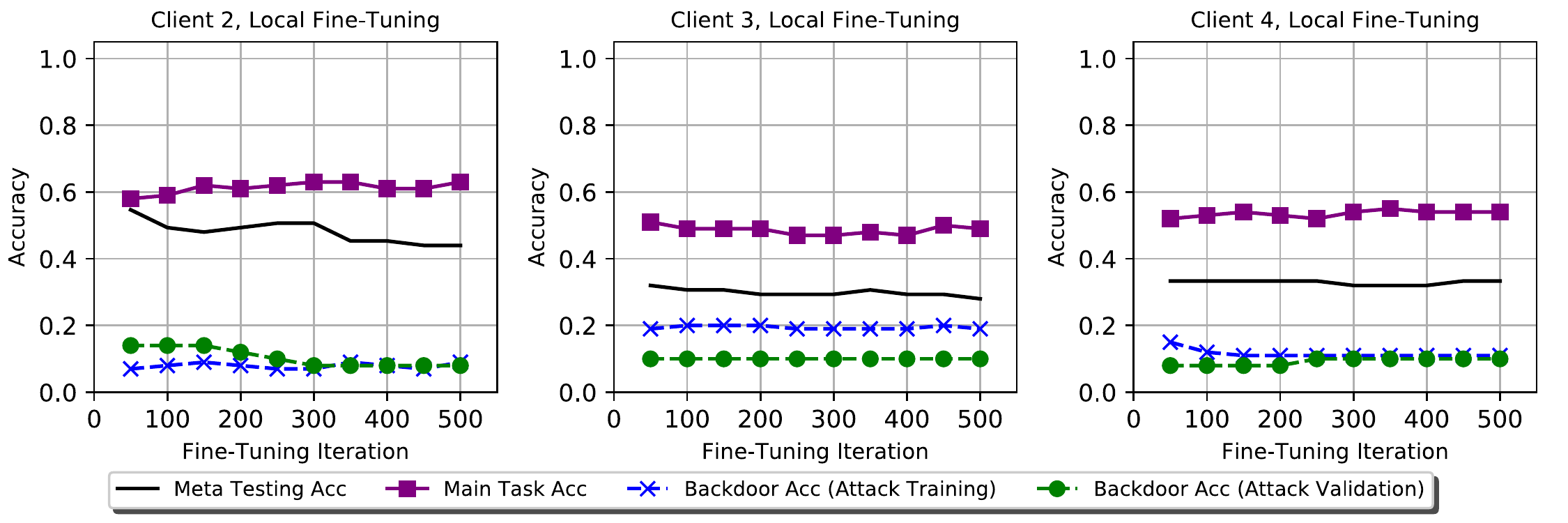}
         \vspace*{-6mm}
         \caption{Backdoor classes used also in benign fine-tuning}
         \label{fig_ap:009_3}
     \end{subfigure}
     % \vspace*{-1mm}
     \caption{Benign local meta-training ($\epsilon = 0.1$, $100$ episodes) and fine-tuning ($\eta = 0.001$) after attacks on mini-ImageNet ($\delta=0.3$)}
     \vspace*{-3mm}
     \label{fig_ap:009}
     \end{minipage}
     \hfill
     \centering
     \begin{minipage}[t]{.495\textwidth}
     \centering
     \begin{subfigure}[hbt!]{\textwidth}
         \centering
         \includegraphics[width=\textwidth]{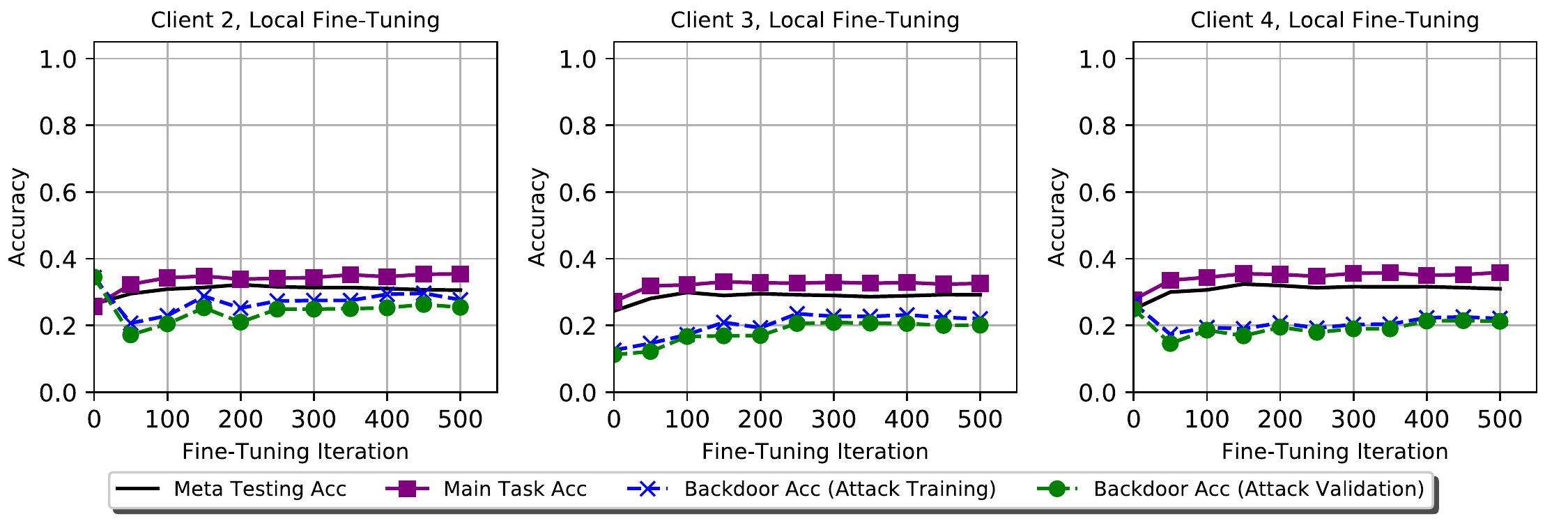}
         \vspace*{-6mm}
         \caption{Backdoor examples not used by benign users}
         \label{fig_ap:009m_1}
     \end{subfigure}
     \hfill
     \vspace*{1mm}
     \begin{subfigure}[hbt!]{\textwidth}
         \centering
         \includegraphics[width=\textwidth]{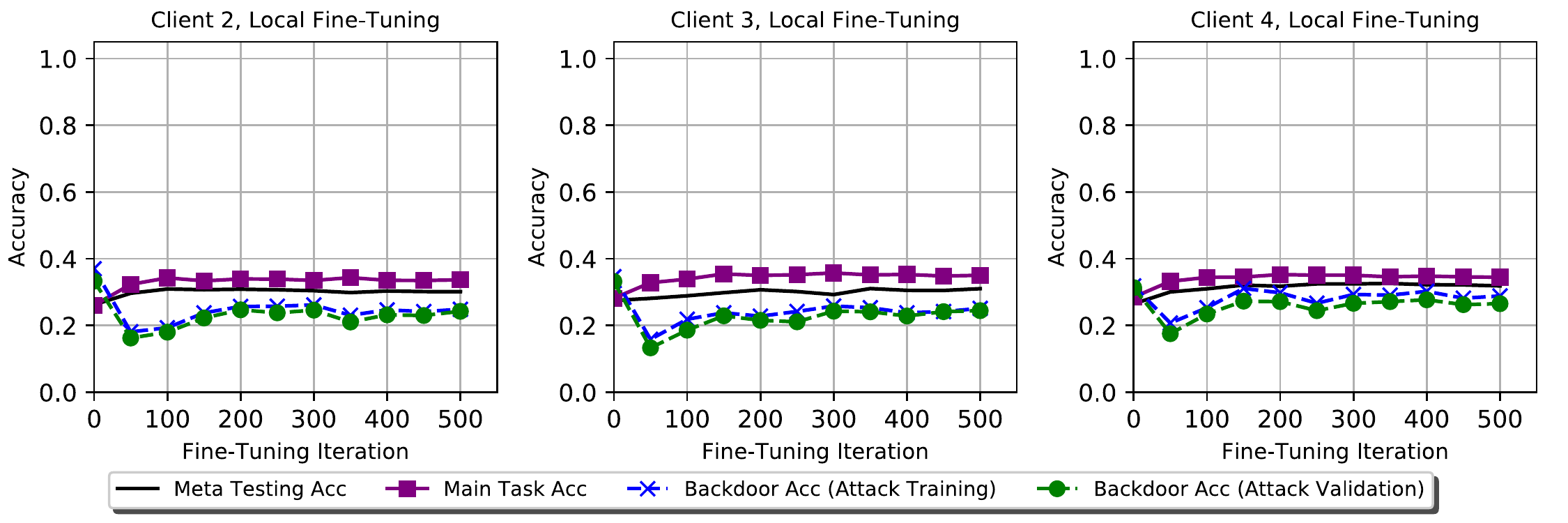}
         \vspace*{-6mm}
         \caption{Backdoor classes used in benign pre-training}
         \label{fig_ap:009m_2}
     \end{subfigure}
     \hfill
     \vspace*{1mm}
     \begin{subfigure}[hbt!]{\textwidth}
         \centering
         \includegraphics[width=\textwidth]{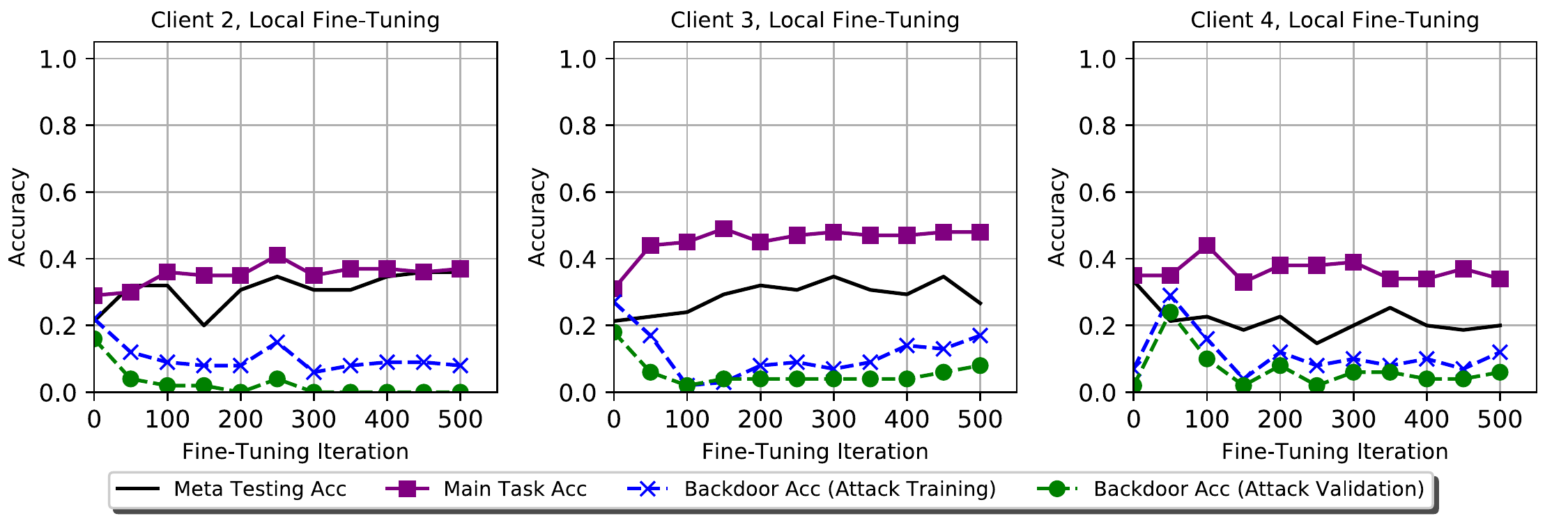}
         \vspace*{-6mm}
         \caption{Backdoor classes used also in benign fine-tuning}
         \label{fig_ap:009m_3}
     \end{subfigure}
     % \vspace*{-1mm}
     \caption{Benign local meta-training ($\epsilon = 0.1$, $100$ episodes) and fine-tuning of matching network ($\eta = 0.001$) after attacks on mini-ImageNet ($\delta=0.3$)}
     \vspace*{-3mm}
     \label{fig_ap:009m}
     \end{minipage}
     % \vspace*{3mm}
\end{figure}

\begin{figure}[t!]
     \centering
     \begin{minipage}[t]{.495\textwidth}
     \centering
     \begin{subfigure}[hbt!]{\textwidth}
         \centering
         \includegraphics[width=\textwidth]{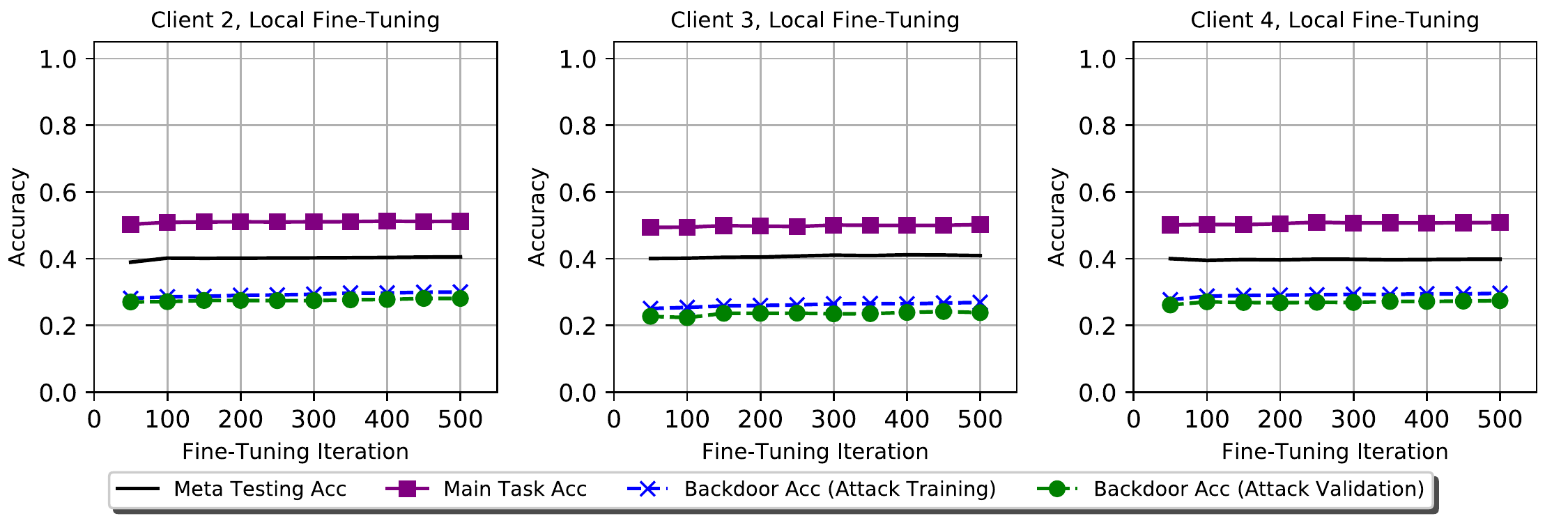}
         \vspace*{-6mm}
         \caption{Backdoor examples not used by benign users}
         \label{fig_ap:010_1}
     \end{subfigure}
     \hfill
     \vspace*{1mm}
     \begin{subfigure}[hbt!]{\textwidth}
         \centering
         \includegraphics[width=\textwidth]{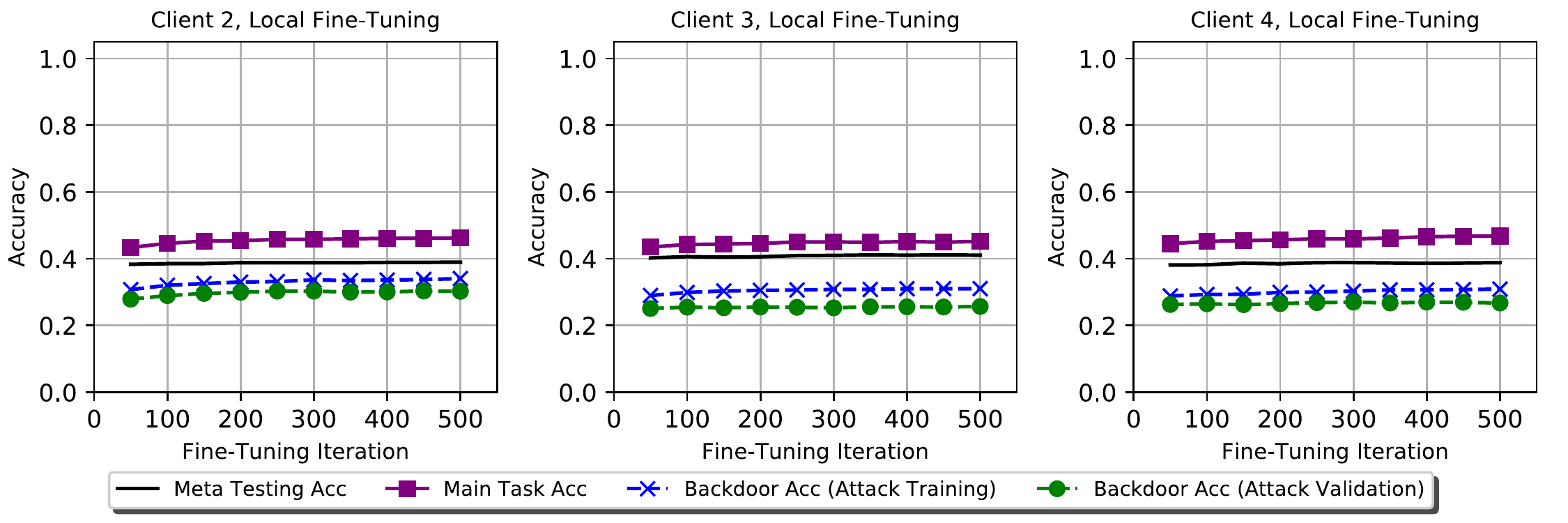}
         \vspace*{-6mm}
         \caption{Backdoor classes used in benign pre-training}
         \label{fig_ap:010_2}
     \end{subfigure}
     \hfill
     \vspace*{1mm}
     \begin{subfigure}[hbt!]{\textwidth}
         \centering
         \includegraphics[width=\textwidth]{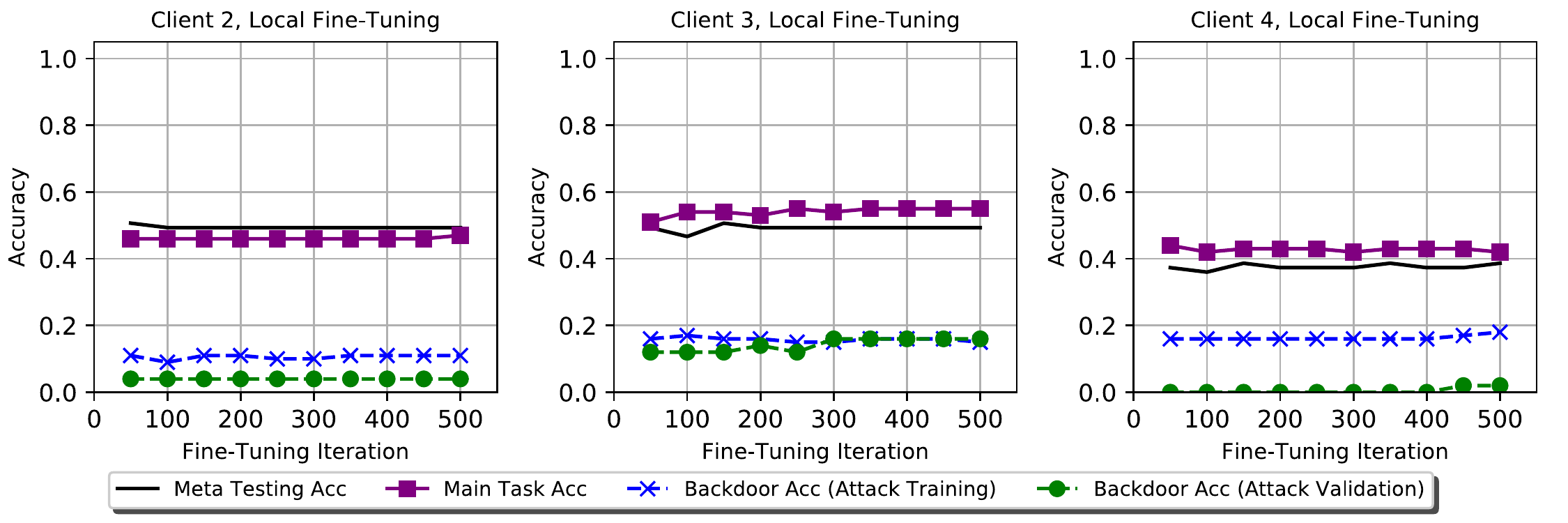}
         \vspace*{-6mm}
         \caption{Backdoor classes used also in benign fine-tuning}
         \label{fig_ap:010_3}
     \end{subfigure}
     % \vspace*{-1mm}
     \caption{Benign local meta-training ($\epsilon = 0.1$, $1000$ episodes) and fine-tuning ($\eta = 0.001$) after attacks on mini-ImageNet ($\delta=0.3$)}
     \vspace*{-3mm}
     \label{fig_ap:010}
     \end{minipage}
     \hfill
     \centering
     \begin{minipage}[t]{.495\textwidth}
     \centering
     \begin{subfigure}[hbt!]{\textwidth}
         \centering
         \includegraphics[width=\textwidth]{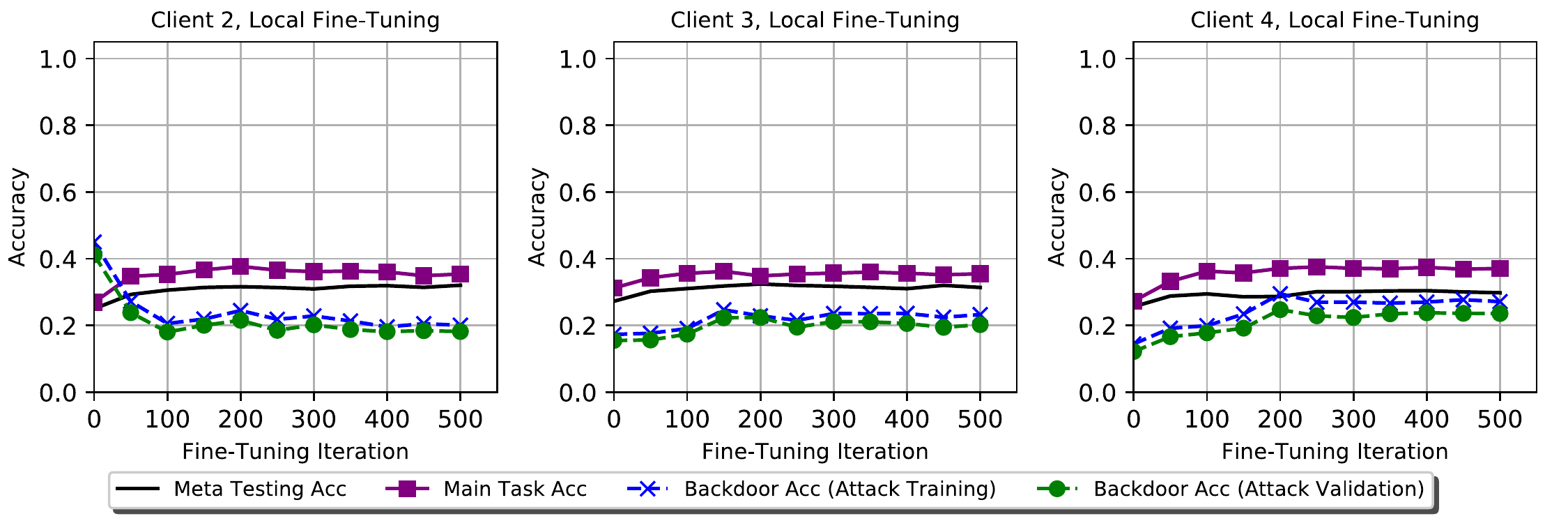}
         \vspace*{-6mm}
         \caption{Backdoor examples not used by benign users}
         \label{fig_ap:010m_1}
     \end{subfigure}
     \hfill
     \vspace*{1mm}
     \begin{subfigure}[hbt!]{\textwidth}
         \centering
         \includegraphics[width=\textwidth]{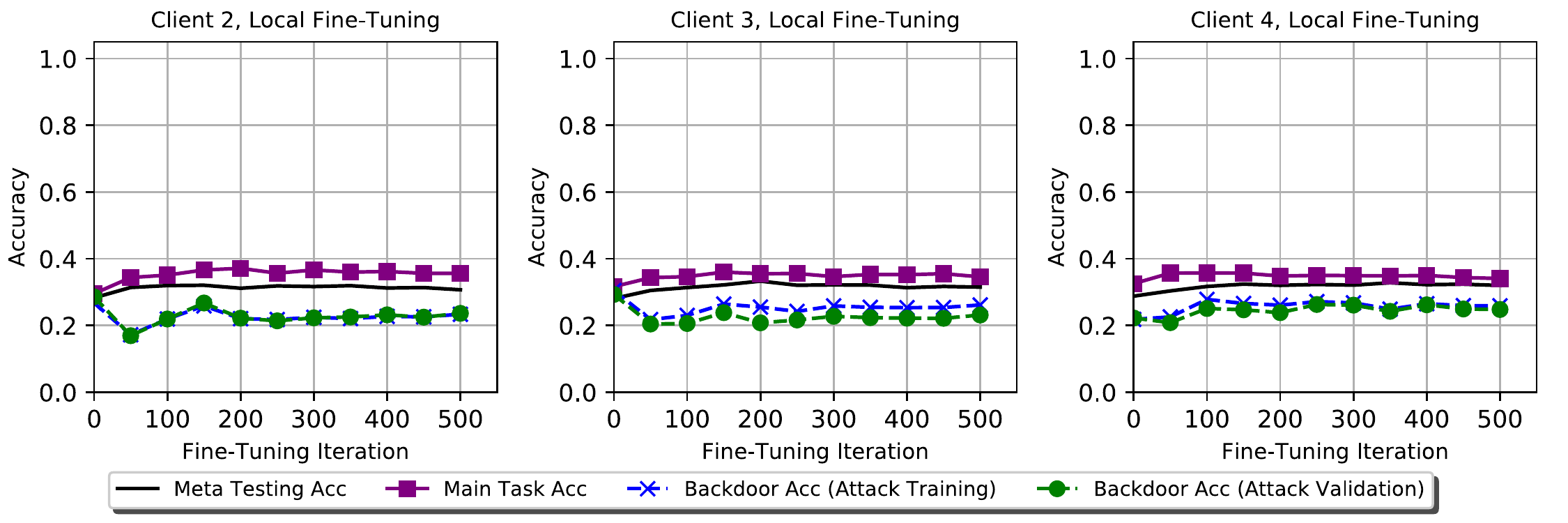}
         \vspace*{-6mm}
         \caption{Backdoor classes used in benign pre-training}
         \label{fig_ap:010m_2}
     \end{subfigure}
     \hfill
     \vspace*{1mm}
     \begin{subfigure}[hbt!]{\textwidth}
         \centering
         \includegraphics[width=\textwidth]{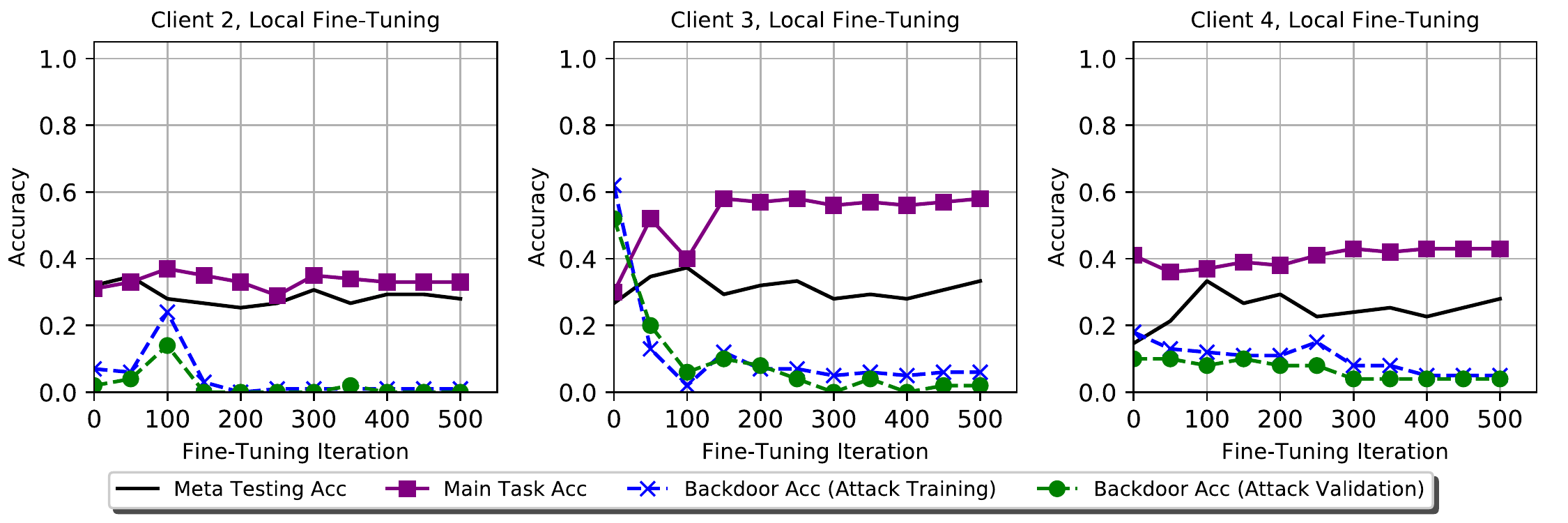}
         \vspace*{-6mm}
         \caption{Backdoor classes used also in benign fine-tuning}
         \label{fig_ap:010m_3}
     \end{subfigure}
     % \vspace*{-1mm}
     \caption{Benign local meta-training ($\epsilon = 0.1$, $1000$ episodes) and fine-tuning of matching network ($\eta = 0.001$) after attacks on mini-ImageNet ($\delta=0.3$)}
     \vspace*{-3mm}
     \label{fig_ap:010m}
     \end{minipage}
     \vspace*{3mm}
\end{figure}

We report results after running $100$ (\cref{fig_ap:009,fig_ap:009m}) or $1000$ (\cref{fig_ap:010,fig_ap:010m}) episodes of additional local meta-training (Reptile) with mini-ImageNet before supervised fine-tuning or matching networks training. 
We note that: (i) results shown in \cref{fig_ap:009,fig_ap:009m,fig_ap:010,fig_ap:010m} are not significantly different from the case without extra local meta-training (\cref{fig_ap:008m} for supervised fine-tuning, and \cref{fig:005m} for fine-tuning of matching networks), suggesting that additional local meta-training does not improve main-task accuracy nor meta-testing accuracy; and (ii) supervised fine-tuning performs similarly to fine-tuning of matching networks, except for \cref{fig_ap:009_1}, in which main-task accuracy and meta-testing accuracy drop to $20\%$ (random guessing over $5$ classes) at the beginning, with high attack accuracy thereafter. This suggests that, by applying random initialization parameters (with additional local training), supervised fine-tuning can behave arbitrarily and may not guarantee removal of backdoor attacks,
while fine-tuning of matching networks performs in a more robust manner.
%Randomness added to the decision logic (weights of the fully connected layer), it may cause unexpected influences to the model through back-propagation and thus, with extra training, may diverge. 
%In contrast, fine-tuning of matching networks is more robust due to its design.

\section{Experimental setup and run time}
% \subsection{Experimental setup and run time}
We implemented federated meta learning using
TensorFlow \cite{tensorflow2015-whitepaper} and Keras \cite{chollet2015keras}. All experiments are performed using 5 virtual machines (VMs) on Google Compute Engine, including 1 federated-learning server and 4 federated-learning clients. Each client VM has 4 Intel Skylake (or later) CPUs and 1 Nvidia Tesla T4 GPU, with 26GB of RAM and Debian 9 OS with CUDA 10.0; each server VM has 1 Intel Skylake (or later) CPU, 3.5GB RAM, and the same version of OS and CUDA.

We report the run time of our code in the following table.

\begin{tabular}{|l|c|c|}
\cline{1-3}
& Omniglot & mini-ImageNet  \\ \cline{1-3}
Pre-training (100k episodes) & $\approx250$ mins & $\approx250$ mins  \\ \cline{1-3}
\begin{tabular}[c]{@{}l@{}}Poisoning training\\ (50k episodes with 50 inner epochs for Omniglot; \\ 150k episodes with 1 inner epoch for mini-ImageNet)\end{tabular} &
  $\approx1250$ mins &
  $\approx450$ mins 
   \\ \cline{1-3}
\begin{tabular}[c]{@{}l@{}}Benign fine-tuning (500 iterations; repeat for 40 tasks \\ and present/absent of backdoor classes)\end{tabular} &
  $\approx130$ mins &
  $\approx130$ mins 
   \\ \cline{1-3}
\begin{tabular}[c]{@{}l@{}}Benign meta-training after attack\\ (50 rounds for Omniglot; 100 rounds for mini-ImageNet)\end{tabular} &
  $\approx80$ mins & 
  $\approx120$ mins 
   \\ \cline{1-3}
\end{tabular}

